\documentclass{article}
\usepackage{ijcai24}
\usepackage{times}
\usepackage{soul}
\usepackage{url}
\usepackage[hidelinks]{hyperref}
\usepackage[utf8]{inputenc}
\usepackage[small]{caption}
\usepackage{graphicx}
\usepackage{amsmath}
\usepackage{amsthm}
\usepackage{amssymb}
\usepackage{booktabs}
\usepackage[linesnumbered,ruled,vlined]{algorithm2e}
\usepackage{algpseudocode}
\usepackage[switch]{lineno}
\usepackage{xcolor}
\usepackage{adjustbox}
\usepackage{bm}
\usepackage{tabularray}
\usepackage{subcaption}
\usepackage{hyperref}

\makeatletter
\def\thmhead@plain#1#2#3{%
  \thmname{#1}\thmnumber{\@ifnotempty{#1}{ }\@upn{#2}}%
  \thmnote{ {\the\thm@notefont#3}}}
\let\thmhead\thmhead@plain
\makeatother

\newtheorem*{definition1}{Definition1}

\newtheorem*{proposition1}{Proposition1}

\title{Learning Pareto Set for Multi-Objective Continuous Robot Control}

\author{
Tianye Shu$^1$
\and
Ke Shang$^{*1,2}$\and
Cheng Gong$^{1,3}$\and
Yang Nan$^1$\And
Hisao Ishibuchi$^{*1}$
\affiliations
$^1$Department of Computer Science and Engineering, Southern University of Science and Technology\\
$^2$National Engineering Laboratory for Big Data System Computing Technology, Shenzhen University\\
$^3$Department of Computer Science, City University of Hong Kong\\
\emails
12132356@mail.sustech.edu.cn,
kshang@foxmail.com,\\
\{12150059, nany\}@mail.sustech.edu.cn,
hisao@sustech.edu.cn
}

\begin{document}
\maketitle
\begin{abstract}
For a control problem with multiple conflicting objectives, there exists a set of Pareto-optimal policies called the Pareto set instead of a single optimal policy. When a multi-objective control problem is continuous and complex, traditional multi-objective reinforcement learning (MORL) algorithms search for many Pareto-optimal deep policies to approximate the Pareto set, which is quite resource-consuming. In this paper, we propose a simple and resource-efficient MORL algorithm that learns a continuous representation of the Pareto set in a high-dimensional policy parameter space using a single hypernet. The learned hypernet can directly generate various well-trained policy networks for different user preferences. We compare our method with two state-of-the-art MORL algorithms on seven multi-objective continuous robot control problems. Experimental results show that our method achieves the best overall performance with the least training parameters. An interesting observation is that the Pareto set is well approximated by a curved line or surface in a high-dimensional parameter space. This observation will provide insight for researchers to design new MORL algorithms.\par
\end{abstract}
{\let\thefootnote\relax\footnotetext{*Corresponding authors.}}
\section{Introduction}
In many real-world control problems, we often have multiple conflicting objectives. For example, in walking robot control, we may need to maximize walking speed and minimize energy consumption. Instead of a single optimal control policy, such a multi-objective control problem has a set of Pareto-optimal control policies called the Pareto set. Each Pareto-optimal policy in the Pareto set corresponds to a different trade-off (i.e., a different user preference) over conflicting objectives. To find these Pareto-optimal policies, multi-objective reinforcement learning (MORL) algorithms are widely used~\cite{roijers2013survey,hayes2022practical,rame2023rewarded}.\par
Many MORL algorithms~\cite{van2014multi,parisi2014Policy,xu2020prediction} search for a finite set of independent policies to approximate the Pareto set. These algorithms have two potential drawbacks when handling continuous multi-objective control problems. One drawback is that a finite number of policies cannot represent the entire Pareto set. Thus, the following undesirable situation can happen: No policy in the obtained finite policy set is close to the user preference. The other drawback is that these methods are memory-consuming when the control problem has many objectives. This is because the number of required policies to approximate the Pareto set increases exponentially as the number of objectives increases. For example, more than 1,000 Pareto-optimal deep policies (represented by deep neural networks) are stored by an evolutionary learning algorithm for a three-objective robot control problem~\cite{xu2020prediction}.\par
One idea to address these drawbacks is to train a single deep neural network to represent the whole Pareto set~\cite{pirotta2015multi,yang2019generalized,basaklar2023pdmorl}. Based on this idea, two different types of methods have been proposed: embedding-based method and manifold-based method. The embedding-based methods learn a generalized policy whose output is conditioned on the input preference~\cite{yang2019generalized,abels2019dynamic,basaklar2023pdmorl}. These algorithms are usually off-policy algorithms with a specifically designed experience replay strategy. The Bellman equation is extended into a multi-objective version by including the user preference~\cite{yang2019generalized,basaklar2023pdmorl}. To improve the sample efficiency, Alegre et al.~\shortcite{lucas2023sample} proposed a novel preference selection strategy based on Generalized Policy Improvement (GPI)~\cite{Barreto2020fast}. The manifold-based methods learn a parametric function to directly approximate the Pareto set as a manifold in the policy parameter space. For example, Pirotta et al.~\shortcite{pirotta2015multi} proposed a gradient-based approach called PMGA. Parisi et al.~\shortcite{parisi2017manifold} improved PMGA by using the importance sampling technique~\cite{owen2000safe}. However, these manifold-based approaches have not yet been applied to deep policies with tens of thousands of parameters. Two difficulties have been pointed out for these manifold-based methods in the literature: (1) The number of required parameters to represent the Pareto set is much more than (e.g., quadratically increases with) the dimensionality of the parameter space~\cite{chen2019meta,basaklar2023pdmorl}, and (2) the Pareto set in the parameter space cannot be efficiently represented by a single continuous policy family~\cite{xu2020prediction}.\par
In this paper, we propose a manifold-based approach called Hyper-MORL to learn the Pareto
sets (i.e., to search for Pareto-optimal deep policies) of multi-objective continuous control problems. Our assumption is that the Pareto set in a high-dimensional parameter space could be well approximated by a continuous manifold in a low-dimensional subspace. This assumption comes from the following property of multi-objective optimization: In general, the Pareto set of an $m$-objective problem is an ($m-1$)-dimensional manifold in an $n$-dimensional parameter space even if $n\gg m$~\cite{hillermeier2001nonlinear}. Based on this assumption, we propose an on-policy MORL algorithm to learn a hypernet that maps a user preference (represented by a direction vector) to a Pareto-optimal policy (represented by a point) in the parameter subspace. We examine our method on several complex multi-objective robot control problems~\cite{xu2020prediction} which are usually solved by deep policies with tens of thousands of parameters. Compared with two state-of-the-art MORL algorithms~\cite{xu2020prediction,basaklar2023pdmorl}, our method shows the best overall performance with the least training parameters. To validate our initial assumption, we investigate the Pareto set learned by our method. We show that the Pareto set of continuous robot control problems can be well approximated by a single continuous policy family.\par
\section{Background}
\label{sec:background}
\subsection{Multi-Objective Markov Decision Process}
A multi-objective RL problem (e.g., multi-objective continuous robot control problem) is usually modeled as a multi-objective Markov decision process (MOMDP). An MOMDP is described by the tuple $\langle S, A, P, \bm{R}, \gamma \rangle$ where $S$, $A$, $P(s'|s,a)$, $\bm{R}$ and $\gamma$ represent state space, action space, probability transition function, reward function and discount factor, respectively. Here the reward function $\bm{R}$: $S\times A\rightarrow \mathbb{R}^m$ returns a reward vector $\bm{r}=(r_1,r_2,...,r_m)$ where $m$ is the number of objectives.\par
For an MOMDP, a policy $\pi(a|s)$ represents the probability of selecting the action $a\in A$ at the state $s\in S$. In each time step $t$, an agent with the policy $\pi$ receives a state $s^{(t)}$ and selects an action $a^{(t)}\sim\pi(a|s^{(t)})$. Then, a reward vector $\bm{r}^{(t+1)}=\bm{R}(s^{(t)},a^{(t)})$ and the next state $s^{(t+1)}\sim P(s|s^{(t)},a^{(t)})$ are given by the MOMDP. In this manner, we can obtain a trajectory $\tau=(s^{(0)},a^{(0)},\bm{r}^{(1)},...,s^{(T-1)},a^{(T-1)},\bm{r}^{(T)})$ where $s^{(0)}$ is the initial state and $T$ is the time horizon. The expected returns $\bm{J}(\pi)=(J_{1}(\pi),J_{2}(\pi),...,J_{m}(\pi))$ is defined as:
\begin{equation}
J_{i}(\pi)=E_{\tau}[\sum_{t=1}^{T}\gamma^{t-1} r^{(t)}_{i}],\:i=1,\:2,\:...,\:m,
\end{equation}
where $r^{(t)}_{i}$ is the $i$-th component of the reward vector $\bm{r}^{(t)}$.\par
Our target is to find a set of Pareto-optimal policies (called Pareto set) to maximize $\bm{J}$. Here the optimality is defined by the Pareto dominance relation as follows.
\begin{definition1}[(Pareto Dominance Relation)] A policy $\pi$ is said to be dominated by another policy $\pi'$ (i.e., $\pi'\succ\pi$) $\textit{iff}$ $\forall i\in \{1,...,m\}$, $J_{i}(\pi')\geq J_{i}(\pi)$ and $\exists j\in \{1,...,m\}$, $ J_{j}(\pi')> J_{j}(\pi)$.
\end{definition1}
In this paper, we assume a policy $\pi(a|s,\theta)$ ($\pi_{\theta}$ for short) is parameterized by ${\theta}\in{\Theta}\subseteq \mathbb{R}^{n}$ where ${\Theta}$ is the parameter space. The Pareto set $PS(\Theta)$ in the parameter space is defined as
\begin{equation}
PS(\Theta)=\{\theta\in\Theta|\nexists \theta'\in\Theta, \pi_{\theta'} \succ\pi_{\theta}\}.
\end{equation}
The image of $PS(\Theta)$ in the objective space is Pareto front.

\subsection{Hypernets}
Hypernets are a type of neural networks that take a context vector as input and generate the parameters for the target neural networks~\cite{chauhan2023brief}. Compared with traditional neural networks, hypernets can be trained in a single-model manner to generate weights for multiple neural networks for solving related tasks~\cite{Oswald2020Continual}. Due to its parameter-efficiency and expressiveness~\cite{galanti2020on}, hypernets have been used in many fields such as reinforcement learning~\cite{sarafian2021Recomposing}. A task-conditioned hypernet is used for generalization across tasks in meta-RL~\cite{jacob2023hypernetworks}. Besides, hypernets have been used in continual RL~\cite{huang2021continual} and zero-shot learning~\cite{2023rezaeihypernetworks} in RL.\par
Recently, hypernets have been used to learn the Pareto set for many multi-objective problems such as multi-objective combinatorial optimization~\cite{lin2022pareto_combinatorial} and multi-task learning~\cite{navon2021learning,lin2019pareto}. In this paper, we use a hypernet to learn the Pareto set $PS(\Theta)$ of an MOMDP.
\section{Proposed Method: Hyper-MORL}\label{sec:method}
To solve a multi-objective RL problem, we propose an efficient manifold-based algorithm called Hyper-MORL.
\subsection{Problem Decomposition}
For multi-objective optimization, decomposition is a commonly-used idea~\cite{zhang2007moea}. A multi-objective RL (MORL) problem can be decomposed into several single-objective subproblems using a preference set $\{\bm{\omega}^{(1)},\bm{\omega}^{(2)},...,\bm{\omega}^{(k)}\}$ and a scalarization function $f$. Each subproblem with preference $\bm{\omega}$ is a single-objective RL problem $SORL(\bm{\omega})$ which aims to maximize a scalar expected return $f(\bm{J},\bm{\omega})$.\par
The scalarization function $f$ has two types: linear and non-linear. The linear scalarization function $f_{LS}(\bm{J},\bm{\omega})=\bm{\omega}^{T}\bm{J}$ is straightforward and widely used in many MORL algorithms~\cite{parisi2014Policy,xu2020prediction}. However, one limitation of $f_{LS}$ is that it can only find a convex hull, which is a subset of the Pareto set~\cite{hayes2022practical}. 
Non-linear scalarization functions can find the whole Pareto set regardless of the Pareto front shape. However, the use of non-linear scalarization functions in RL is difficult since the assumed additive property in the Bellman equation does not hold for the non-linearly scalarized returns~\cite{roijers2018multi,hayes2022practical}.\par
In this paper, we consider the linear scalarization function since Lu et al.~\shortcite{lu2023multiobjective} theoretically revealed that a MORL problem is convex (i.e., the range of the value function is convex) when all stationary policies are considered. Given a preference space $\Omega=\{\bm{\omega}|\bm{\omega}\in (\{0\}\cup\mathbb{R}^{+})^{m},\sum_{i=1}^{m}\omega_{i}=1\}$, the following proposition holds.\par
\begin{proposition1}[~\cite{lu2023multiobjective}]
For a convex MORL problem, a policy $\pi\in PS(\Theta)$ iff $\exists \bm{\omega}\in\Omega$, $\pi$ is optimal for the problem $SORL(\bm{\omega})$ which maximizes $\bm{\omega}^{T}\bm{J}(\pi)$.
\end{proposition1}
Proposition 1 says that we can find the whole Pareto set by solving the problem $SORL(\bm{\omega})$ for each preference $\bm{\omega}\in\Omega$.
\subsection{Pareto Set Representation via Hypernet}\label{sec:representation}
\begin{figure}[!h]
    \centering
\includegraphics[width=\linewidth]{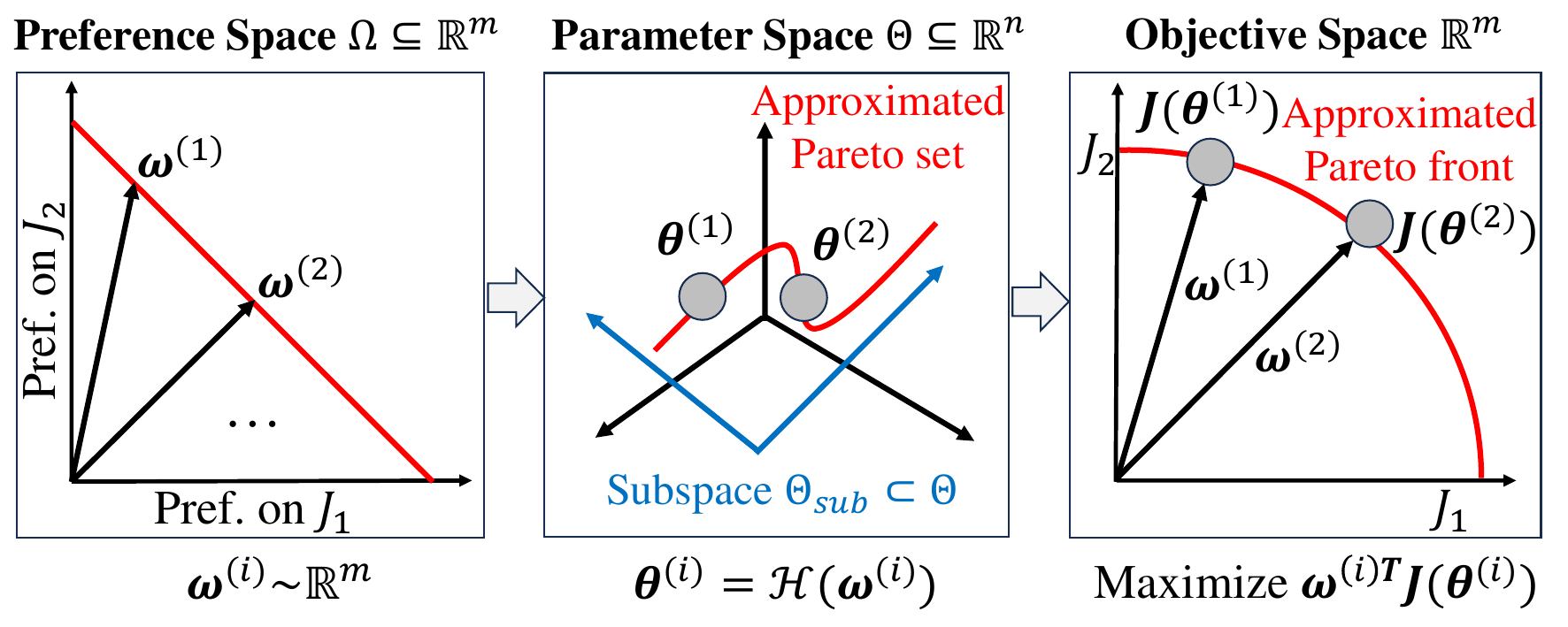}
    \caption{Basic idea of Hyper-MORL to approximate the Pareto set in the $n$-dimensional parameter space.}
    \label{fig:illustration}
\end{figure}

The basic idea of Hyper-MORL is shown in Figure~\ref{fig:illustration}. To represent the Pareto set, we train a model to map each preference $\bm{\omega}\in\Omega$ to the corresponding Pareto-optimal policy parameter $\bm{\theta}\in\Theta$ which maximizes $\bm{\omega}^{T}\bm{J}(\bm{\theta})$. To reduce the number of the required parameters, we consider a reduced $d$-dimensional subspace $\Theta_{sub}$ instead of the original $n$-dimensional parameter space $\Theta$ to represent the Pareto set ($d\ll n$). The value of $d$ is specified as 10 in this paper (and some other values are also examined for sensitivity analysis). A hypernet $\mathcal{H}_{\bm{\varphi}}$: $\Omega\rightarrow\Theta$ is defined as:
\begin{equation}
\mathcal{H}_{\bm{\varphi}}(\bm{\omega}) = \bm{W}f_{\bm{\mu}}(\bm{\omega})+\bm{b}.
\label{eq:hypernetwork}
\end{equation}
Here $f_{\bm{\mu}}$ maps the preference $\bm{\omega}$ to a $d$-dimensional vector. The parameters $\bm{W}$ and $\bm{b}$ are an $n\times d$ matrix and an $n$-dimensional vector, respectively. The row vectors of $\bm{W}$ spans a $d$-dimensional parameter subspace $\Theta_{sub}$. The vector $f_{\bm{\mu}}(\bm{\omega})$ is linearly transformed from $\Theta_{sub}$ to $\Theta$. The learning parameters $\bm{\varphi}$ of the hypernet $\mathcal{H}_{\bm{\varphi}}$ are $\{\bm{W},\bm{\mu},\bm{b}\}$. A detailed illustration of $\mathcal{H}_{\bm{\varphi}}$ is included in Appendix\footnote{\href{https://github.com/HisaoLabSUSTC/Hyper-MORL/blob/main/Appendix.pdf}{https://github.com/HisaoLabSUSTC/\text{Hyper-MORL}/blob/\\main/Appendix.pdf}} B.
\subsection{Pareto Set Learning via Policy Gradient}
To efficiently learn the Pareto set, a gradient-based algorithm is proposed for Hyper-MORL. As shown in Algorithm~\ref{alg:Learning}, the learning algorithm contains warm-up stage and Pareto set learning stage. As shown in lines 1-2, $\alpha\times 100\%$ computation load is assigned to the warm-up stage where $\alpha$ is a parameter. In the following part, we will describe these two stages.
\begin{algorithm}[!t]
\caption{Learning Pareto Set via Policy Gradient}
\label{alg:Learning}
\SetKwInOut{Input}{Input}
\Input{Hypernet $\mathcal{H}_{\bm{\varphi}}$: $\Omega\rightarrow\Theta$ with parameters $\bm{\varphi}=\{\bm{W},\bm{\mu},\bm{b}\}$, the number of sampled preferences $K$, the number of available environment steps $T$, parameter $\alpha$}
$G_{W} = \lfloor\frac{\alpha T}{T_{tra}}\rfloor$\;
$G_{PSL} = \lfloor\frac{(1-\alpha)T}{KT_{tra}}\rfloor$;\textcolor{teal}{\,\,\,// $T_{tra}$ is the number of consumed environment steps to collect a trajectory}\\
\textcolor{teal}{// Warm-up Stage}\\
$\bm{W}\leftarrow\bm{0}$\;
Set $\bm{\mu}$ and $\bm{b}$ by an arbitrary initialization scheme\;
$\bm{\omega}=(\frac{1}{m},...,\frac{1}{m})$\;
\For{$g\leftarrow 1$ \KwTo $G_{W}$}{
Run policy $\pi_{\bm{b}}$ in environment and collect a trajectory $\tau$\;
Calculate gradients
$\bm{g}\leftarrow\nabla_{\bm{b}}[\bm{\omega}^{T}\bm{J}(\bm{b})]$\;
Update parameters $\bm{b}\leftarrow\textbf{ADAM}(\bm{b},\bm{g})$\;
}
Reset exploration-related parameters in $\bm{b}$\;
\textcolor{teal}{// Pareto Set Learning Stage}\\
\For{$g\leftarrow 1$ \KwTo $G_{PSL}$}{
Sample $K$ preferences $\bm{\omega}^{(1)}$, ..., $\bm{\omega}^{(K)}$ from $\Omega$\;
\For{$i\leftarrow 1$ \KwTo $K$}{
Run policy $\pi_{\mathcal{H}_{\bm{\varphi}}(\bm{\omega}^{(i)})}$ in environment and collect a trajectory $\tau^{(i)}$\;
}
\For{$i\leftarrow 1$ \KwTo $K$}{
Calculate gradients $\bm{g}^{(i)}\leftarrow\nabla_{\bm{\varphi}}[\bm{\omega}^{(i)T}\bm{J}(\mathcal{H}_{\bm{\varphi}}(\bm{\omega}^{(i)}))]$ 
}
Update parameters $\bm{\varphi}\leftarrow\textbf{ADAM}(\bm{\varphi},\frac{1}{K}\sum_{i=1}^{K}\bm{g}^{(i)})$\;
}

\end{algorithm}
\subsubsection{Warm-up Stage}
The warm-up stage is to find a good initial policy close to the Pareto front. By starting from a good initial policy instead of a random policy, the Pareto set learning stage can be accelerated. In Algorithm~\ref{alg:Learning}, the warm-up stage is in lines 3-11. In lines 4-5, we use the Bias-HyperInit method~\cite{jacob2023hypernetworks} to initialize the hypernet parameters. In lines 6-10, we update parameters $\bm{b}$  to optimize all objectives simultaneously with the preference $(\frac{1}{m},...,\frac{1}{m})$ ($m$ is the number of objectives). The multi-objective policy gradient method~\cite{xu2020prediction} is used as follows:
\begin{equation}
\begin{split}
\nabla_{\bm{\theta}}&\left[\bm{\omega}^{T}\bm{J}(\bm{\theta})\right]=\sum_{i=1}^{m}\left[\omega_{i}\nabla_{\bm{\theta}}J_{i}(\bm{\theta})\right]\\
&=\mathbb{E}\left[\sum_{t=0}^{T}\left[\bm{\omega}\bm{A}(s_{t},a_{t})\nabla_{\bm{\theta}}\log{\pi_{\bm{\theta}}(a_{t}|s_{t})}\right]\right], 
\end{split}
\label{eq:policy_gradient}
\end{equation}where $\bm{A}$ is the advantage function estimated from a sampled trajectory $\tau$.\par
After $G_{W}$ iterations, we obtain a policy $\pi_{\bm{b}}$ from the preference $(\frac{1}{m}, ..., \frac{1}{m})$. In our implementation, the policy $\pi_{\bm{b}}$ samples an action from a normal distribution. Some values in the $n$-dimensional vector $\bm{b}$ represent the standard deviation of the output action. In line 11, the exploration-related parameters (i.e., the standard deviation of the output action) are reset as one. Since other values in $\bm{b}$ are not changed, the mean of the output action is preserved. The reset step is to prevent the output action selection from being almost deterministic (i.e., standard deviation from being almost zero) which severely degrades the exploration ability in the Pareto set learning stage. After the warm-up stage, $\bm{b}$ becomes closer to the Pareto set, and $\bm{W}$ remains $\bm{0}$. From Eq. (\ref{eq:hypernetwork}), the hypernet $\mathcal{H}_{\bm{\varphi}}(\bm{\omega})=\bm{b}$ for $\forall \bm{\omega}\in \Omega$. This means that the initial values of $\bm{b}$ are shared among all subproblems.\par
\subsubsection{Pareto Set Learning Stage}
Our task is to find the optimal parameters $\bm{\varphi}$ (i.e., $\bm{W}$, $\bm{\mu}$ and $\bm{b}$) which maximizes $\bm{\omega}^{T}\bm{J}(\mathcal{H}_{\bm{\varphi}}(\bm{\omega}))$ for each $\bm{\omega}\in\Omega$. The objective function can be written as:
\begin{equation}
\mathbb{E}_{\bm{\omega}\sim\Omega}\left[\bm{\omega}^{T}\bm{J}(\mathcal{H}_{\bm{\varphi}}(\bm{\omega}))\right].
\label{eq:objective_function}
\end{equation}\par
It is difficult to directly optimize the parameters $\bm{\varphi}$ to maximize the exception in Eq. (\ref{eq:objective_function}). Thus, we use Monte Carlo method to sample the preferences, and use the policy gradient method to optimize the parameters $\bm{\varphi}$. In Algorithm~\ref{alg:Learning}, lines 13-19 show the Pareto set learning stage. In each iteration, $K$ preferences are randomly sampled from the preference space $\Omega$. Then, we obtain $K$ policies by inputting these $K$ preferences into the hypernet. In lines 15-16, each policy interacts with the environment and generates a trajectory. In lines 17-18, the gradient $\bm{g}^{(i)}$ is calculated for each preference based on the corresponding trajectory. Then, the parameters $\bm{\varphi}$ are updated based on the average gradient over all sampled preferences (i.e., $\frac{1}{K}\sum_{i=1}^{K}\bm{g}^{(i)}$).\par
The remaining question is how to calculate the gradient $\bm{g}^{(i)}$ in line 18. By the chain rule, we have 
\begin{equation}
\resizebox{0.9\linewidth}{!}{$
\nabla_{\bm{\varphi}}\left[\bm{\omega}^{T}\bm{J}(\mathcal{H}_{\bm{\varphi}}(\bm{\omega}))\right]= 
    \nabla_{\bm{\varphi}}\left[ \nabla_{\bm{\theta}}\left[\bm{\omega}^{T}\bm{J}(\bm{\theta})\right]\cdot\mathcal{H}_{\bm{\varphi}}(\bm{\omega})\right]$,}
    \label{eq:hypernet_gradient}
\end{equation}
where $\bm{\theta}=\mathcal{H}_{\bm{\varphi}}(\bm{\omega})$ and $\cdot$ means dot product. As we can see, the term $\nabla_{\bm{\theta}}\left[\bm{\omega}^{T}\bm{J}(\bm{\theta})\right]$ can be calculated by Eq. (\ref{eq:policy_gradient}). Thus, the gradient $\bm{g}^{(i)}$ can be calculated by Eq. (\ref{eq:policy_gradient}) and Eq. (\ref{eq:hypernet_gradient}). After the Pareto set learning stage, the whole Pareto set $PS(\Theta)$ is obtained from the hypernet $\mathcal{H}_{\bm{\varphi}}$. 


\section{Experiments}\label{sec:experiment}
\subsection{Benchmark Problems}
To test the performance of Hyper-MORL, we use seven problems in Table~\ref{tab:benchmark} from a multi-objective robot control benchmark suite~\cite{xu2020prediction}. For more details, please refer to Appendix A.
 \begin{table}[!h]
    \centering
    \begin{adjustbox}{width=0.85\linewidth}
    \begin{tblr}{cccc}\hline
      Problem & $m$ & State space & Action space \\ \hline
     MO-Swimmer-v2     & 2  &$\mathcal{S}\subseteq \mathbb{R}^{8}$  &  $\mathcal{A}\subseteq \mathbb{R}^{2}$ \\
     MO-HalfCheetah-v2 & 2  &$\mathcal{S}\subseteq \mathbb{R}^{17}$ & $\mathcal{A}\subseteq \mathbb{R}^{6}$ \\
     MO-Walker2d-v2    & 2  &$\mathcal{S}\subseteq \mathbb{R}^{17}$ & $\mathcal{A}\subseteq \mathbb{R}^{6}$\\
     MO-Ant-v2         & 2  &$\mathcal{S}\subseteq \mathbb{R}^{27}$ &$\mathcal{A}\subseteq \mathbb{R}^{8}$\\
     MO-Hopper-v2      & 2  &$\mathcal{S}\subseteq \mathbb{R}^{11}$ &$\mathcal{A}\subseteq \mathbb{R}^{3}$\\
      MO-Humanoid-v2    & 2 &$\mathcal{S}\subseteq \mathbb{R}^{376}$ &$\mathcal{A}\subseteq \mathbb{R}^{17}$\\
     MO-Hopper-v3      & 3  &$\mathcal{S}\subseteq \mathbb{R}^{11}$ &$\mathcal{A}\subseteq \mathbb{R}^{3}$
    \\\hline
    \end{tblr}
    \end{adjustbox}
    \caption{Seven test problems in a multi-objective robot control benchmark~\protect\cite{xu2020prediction}.}
    \label{tab:benchmark}
\end{table}
\subsection{Compared MORL Algorithms}
We compare the proposed Hyper-MORL algorithm with the following two state-of-the-art MORL algorithms.
\subsubsection{PG-MORL~\protect\cite{xu2020prediction}}
Prediction-guided MORL (i.e., PG-MORL) algorithm is an evolutionary learning algorithm. In each generation, PG-MORL selects a set of policies from the current population based on a prediction model, and optimizes them with a multi-objective policy gradient method. All Pareto-optimal policies are stored in an unbounded external archive. We use the default settings in the original paper for PG-MORL\footnote{\href{https://github.com/mit-gfx/PGMORL}{https://github.com/mit-gfx/PGMORL}}.
\subsubsection{PD-MORL~\protect\cite{basaklar2023pdmorl}} Preference-driven MORL (i.e., PD-MORL) algorithm learns a generalized policy by combining the preference into the Bellman equation. We use the default settings in the original paper for PD-MORL\footnote{\href{https://github.com/tbasaklar/PDMORL-Preference-Driven-Multi-Objective-Reinforcement-Learning-Algorithm}{https://github.com/tbasaklar/PDMORL-Preference-Driven-Multi-Objective-Reinforcement-Learning-Algorithm}}. Note that the number of environment steps used by PD-MORL is smaller than that used by PG-MORL as shown in Table~\ref{tab:termination_condition}. However, PD-MORL needs more computation time than PG-MORL, which we will discuss in Section~\ref{sec:performance_cp}. We find it impractical for PD-MORL to use the same number of environment steps as in PG-MORL. 
\begin{table}[!t]
    \centering
    \begin{adjustbox}{width=0.98\linewidth}
    \begin{tblr}
    {width=1.0\linewidth,
    cell{1}{1-4} = {c},
    cell{2-8}{1} = {c},
    cell{2-8}{2-4} = {r}}
    \hline
    Problem      &  Hyper-MORL      & PG-MORL          & PD-MORL          \\\hline
    MO-Swimmer-v2    &  $1.2\times10^7$ & $1.2\times10^7$ & $1\times10^7$ \\
    MO-Walker2d-v2     &  $3\times10^7$ & $3\times10^7$ & $1\times10^7$ \\
    MO-HalfCheetah-v2&  $3\times10^7$ & $3\times10^7$ & $1\times10^7$ \\
    MO-Ant-v2        &  $4.8\times10^7$ & $4.8\times10^7$ & $1\times10^7$ \\
    MO-Hopper-v2     &  $4.8\times10^7$ & $4.8\times10^7$ & $1\times10^7$ \\
    MO-Humanoid-v2   &   $1.2\times10^8$ & $1.2\times10^8$ & $1\times10^7$ \\
    MO-Hopper-v3     &  $1.2\times10^8$ & $1.2\times10^8$ & $1\times10^7$ \\\hline
    \end{tblr}
    \end{adjustbox}
    \caption{The number of environment steps used by each algorithm for each problem. PG-MORL and PD-MORL follow the settings in their original papers~\protect\cite{xu2020prediction,basaklar2023pdmorl}.}
    \label{tab:termination_condition}
\end{table}
\subsubsection{Hyper-MORL (Our Algorithm)}
In our algorithm, we use the same neural network architecture as in PG-MORL. Thus, the parameter spaces of Hyper-MORL and PG-MORL are the same. For a fair comparison, Hyper-MORL uses the same policy gradient method (i.e., PPO~\cite{schulman2017proximal}) as in PG-MORL. The optimizer is ADAM with learning rate $\eta = 5\times10^{-5}$. In Hyper-MORL, the parameters $\alpha$ and $d$ are set as 0.15 (i.e., 15\%) and 10, respectively. The number of sampled preferences $K$ is set as 6 and 15 for two-objective and three-objective problems, respectively. For more details, please refer to Appendix C. All codes of Hyper-MORL are available from \url{https://github.com/HisaoLabSUSTC/Hyper-MORL}.
\begin{table*}[!t]
    \centering
\begin{tblr}{
    cell{1}{1-4} = {c},
    cell{2-9}{1} = {c},
    cell{2-8}{2-4} = {r},
    cell{9}{2-4} = {c},
    }\hline
   Algorithm  &Hyper-MORL (Ours) &PG-MORL  & PD-MORL\\\hline
MO-Swimmer-v2 &2.88$\pm$0.44 $\times$ $10^{4}$ (2)&2.60$\pm$0.71 $\times$ $10^{4}$ (3)&$\bm{3.13\pm0.14\times 10^{4} (1)}$\\
MO-HalfCheetah-v2&5.53$\pm$0.10$\times$$10^{6}$ (3)&5.75$\pm$0.02$\times$$10^{6}$ (2)&$\bm{5.89\pm0.01\times10^{6} (1)}$\\
MO-Walker2d-v2&$\bm{5.37\pm0.35\times10^{6} (1)}$&4.41$\pm$0.74$\times$$10^{6}$ (2)&5.08$\pm$0.23$\times$$10^{6}$ (3)\\
MO-Ant-v2&$\bm{7.49\pm0.16\times10^{6} (1)}$&5.79$\pm$0.25$\times$$10^{6}$ (3)&7.05$\pm$1.13$\times$$10^{6}$ (2)\\
MO-Hopper-v2&$\bm{2.05\pm0.06\times10^{7} (1)}$&1.96$\pm$0.17$\times$$10^{7}$ (2)&5.84$\pm$8.12$\times$$10^{6}$ (3)\\
MO-Humanoid-v2&4.32$\pm$0.42$\times$$10^{7}$ (2)&$\bm{4.69\pm0.29\times10^{7} (1)} $&1.66$\pm$0.60$\times$$10^{7}$ (3)\\
MO-Hopper-v3&$\bm{3.58\pm0.25\times10^{10} (1)}$&3.31$\pm$0.23$\times$$10^{10}$ (2)&1.14$\pm$1.24$\times$$10^{10}$ (3)\\\hline
Average Rank&$\bm{1.57}$ &2.29&2.14 \\\hline
\end{tblr}
 \caption{HV-based performance comparison of Hyper-MORL with two state-of-the-art algorithms on a multi-objective continuous robot control benchmark~\protect\cite{xu2020prediction}. The reference point for HV calculation is set as (0, ..., 0). The average values and standard deviations over nine runs are reported. The rank of each method is shown in the parenthesis, and a small value means a better rank.}
         \label{tab:HV}
\end{table*}
\begin{figure*}[!t]
\begin{subfigure}[c]{0.24\textwidth}
    \includegraphics[width=\textwidth]{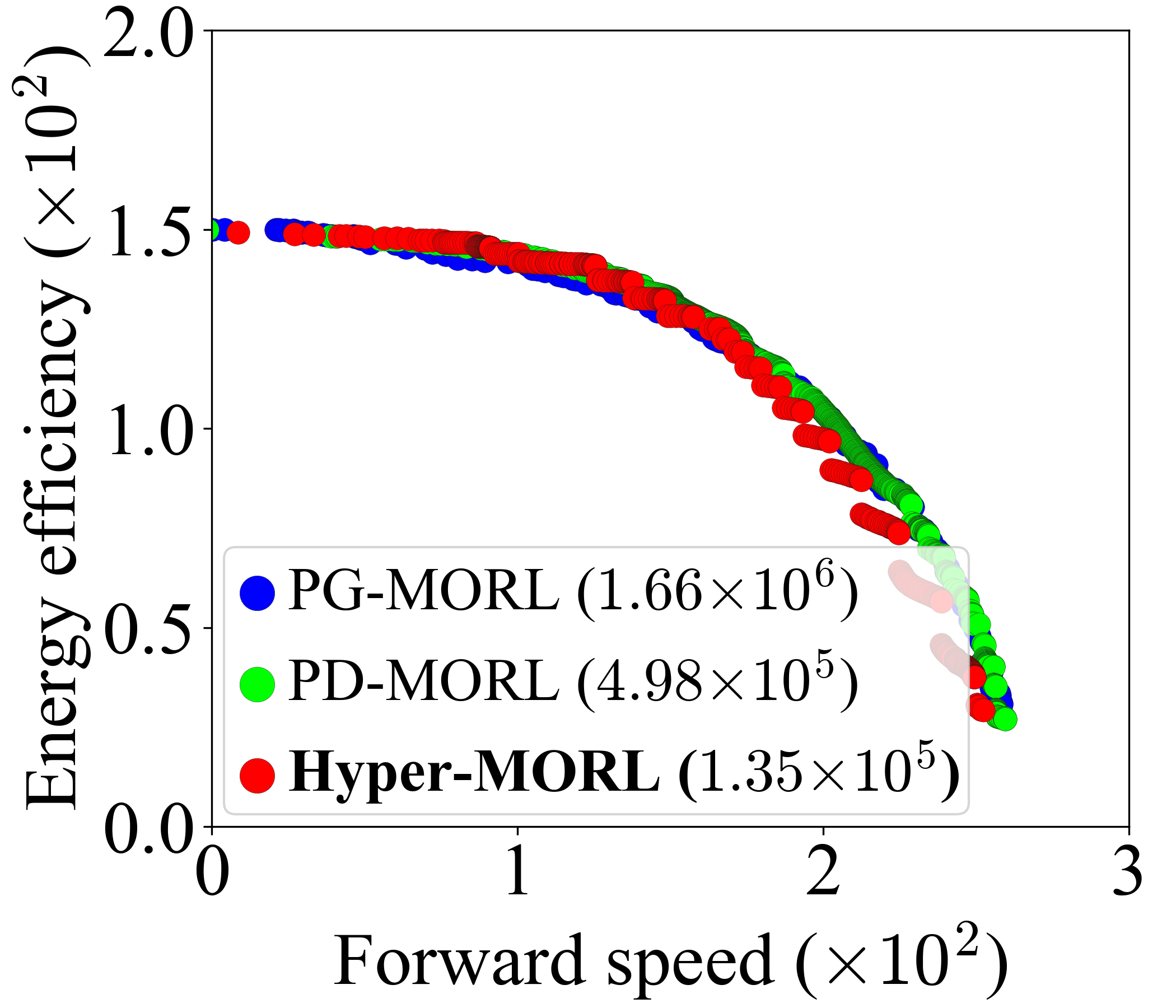}
    \caption{MO-Swimmer-v2}
\end{subfigure}
\begin{subfigure}[c]{0.24\textwidth}
    \includegraphics[width=\textwidth]{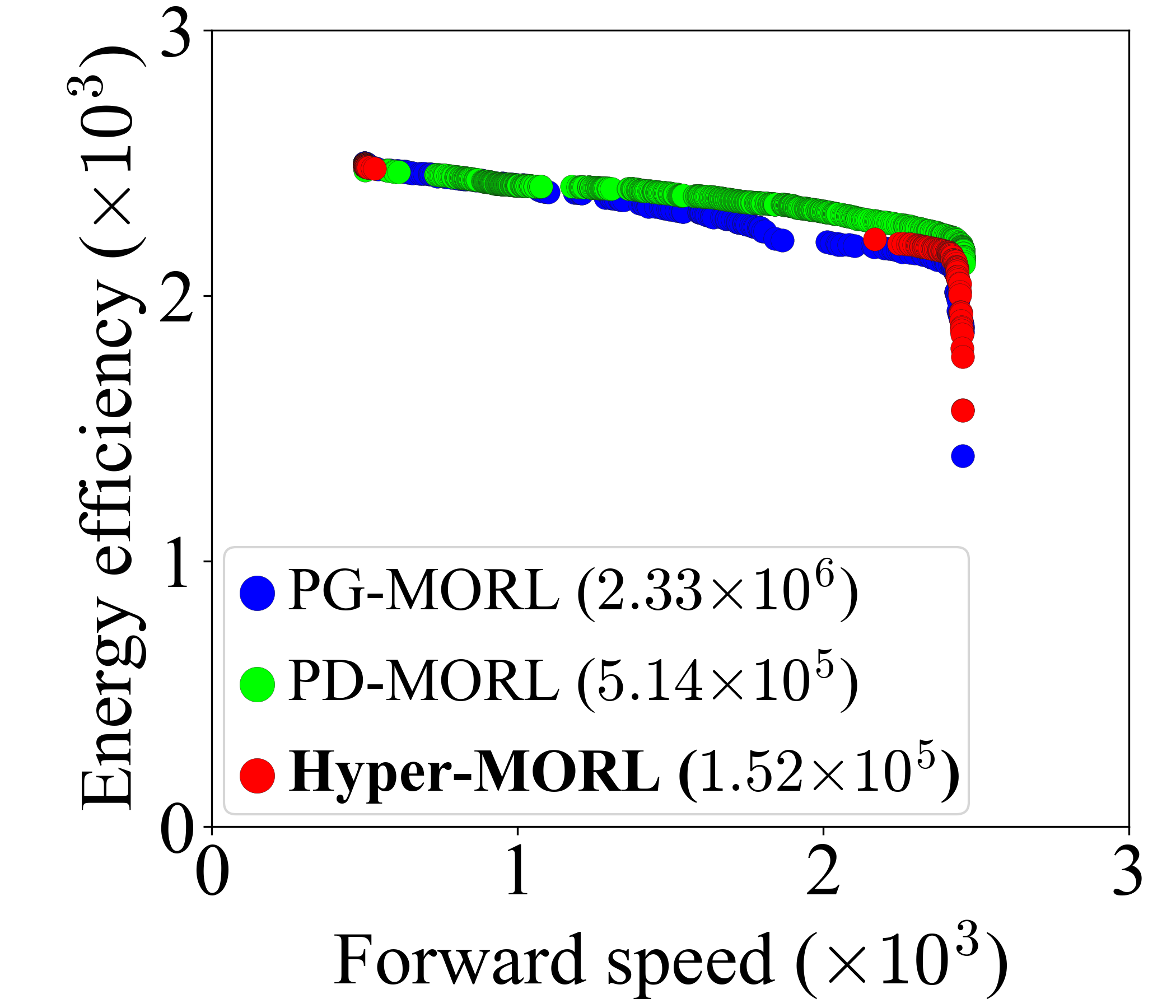}
    \caption{MO-HalfCheetah-v2}
\end{subfigure}
\begin{subfigure}[c]{0.24\textwidth}
    \includegraphics[width=\textwidth]{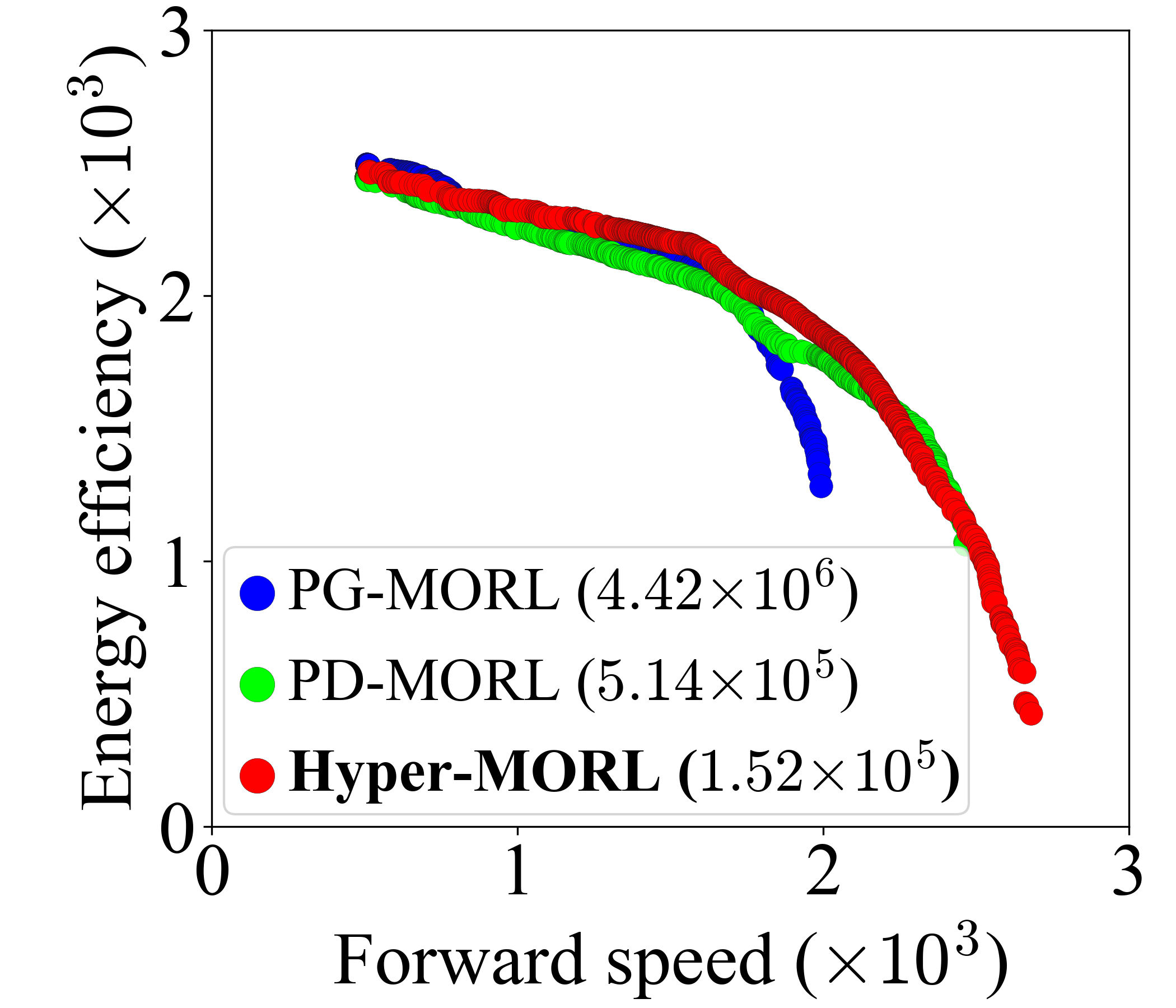}
    \caption{MO-Walker2d-v2}
\end{subfigure}
\begin{subfigure}[c]{0.24\textwidth}
    \includegraphics[width=\textwidth]{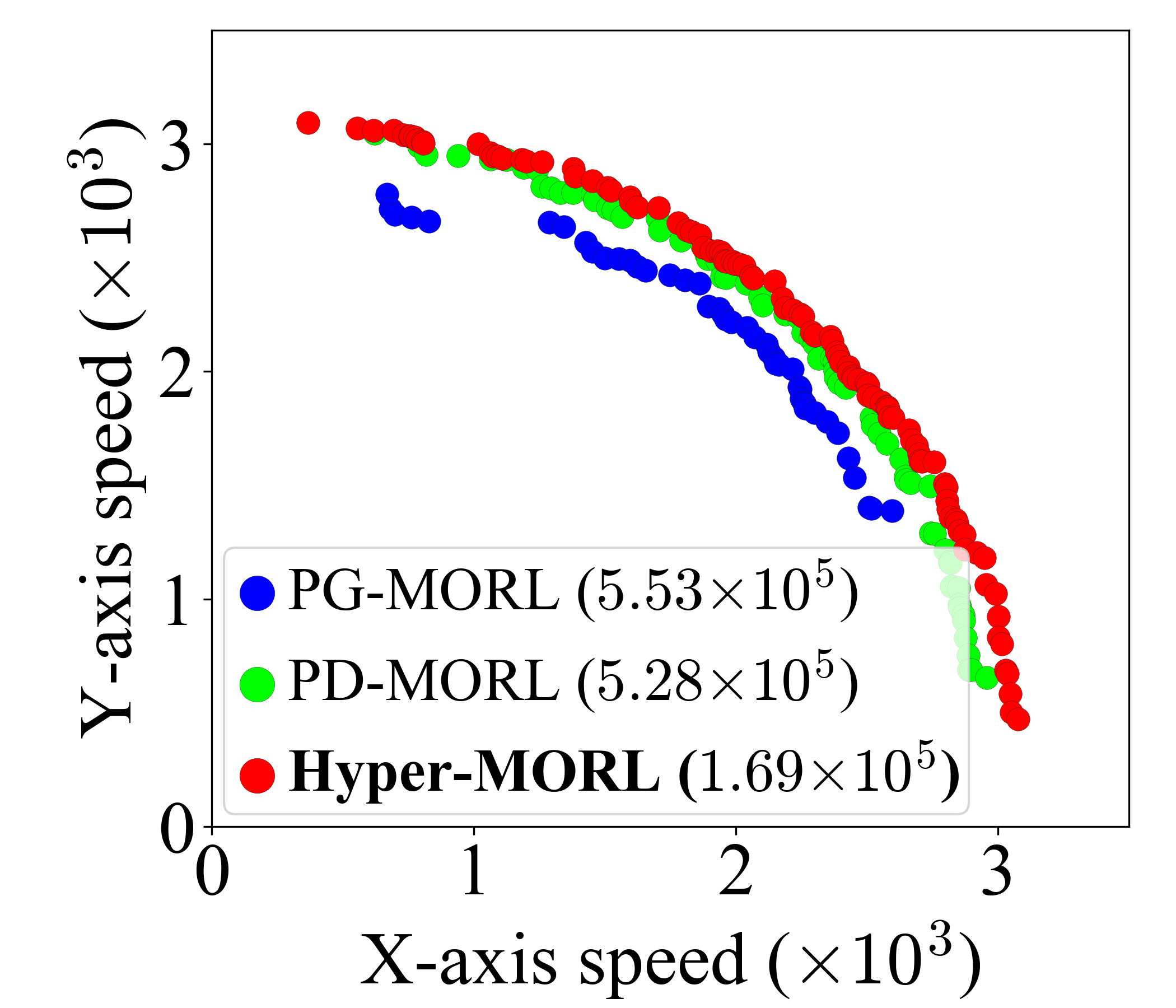}
    \caption{MO-Ant-v2}
\end{subfigure}

\begin{subfigure}[c]{0.24\textwidth}
    \includegraphics[width=\textwidth]{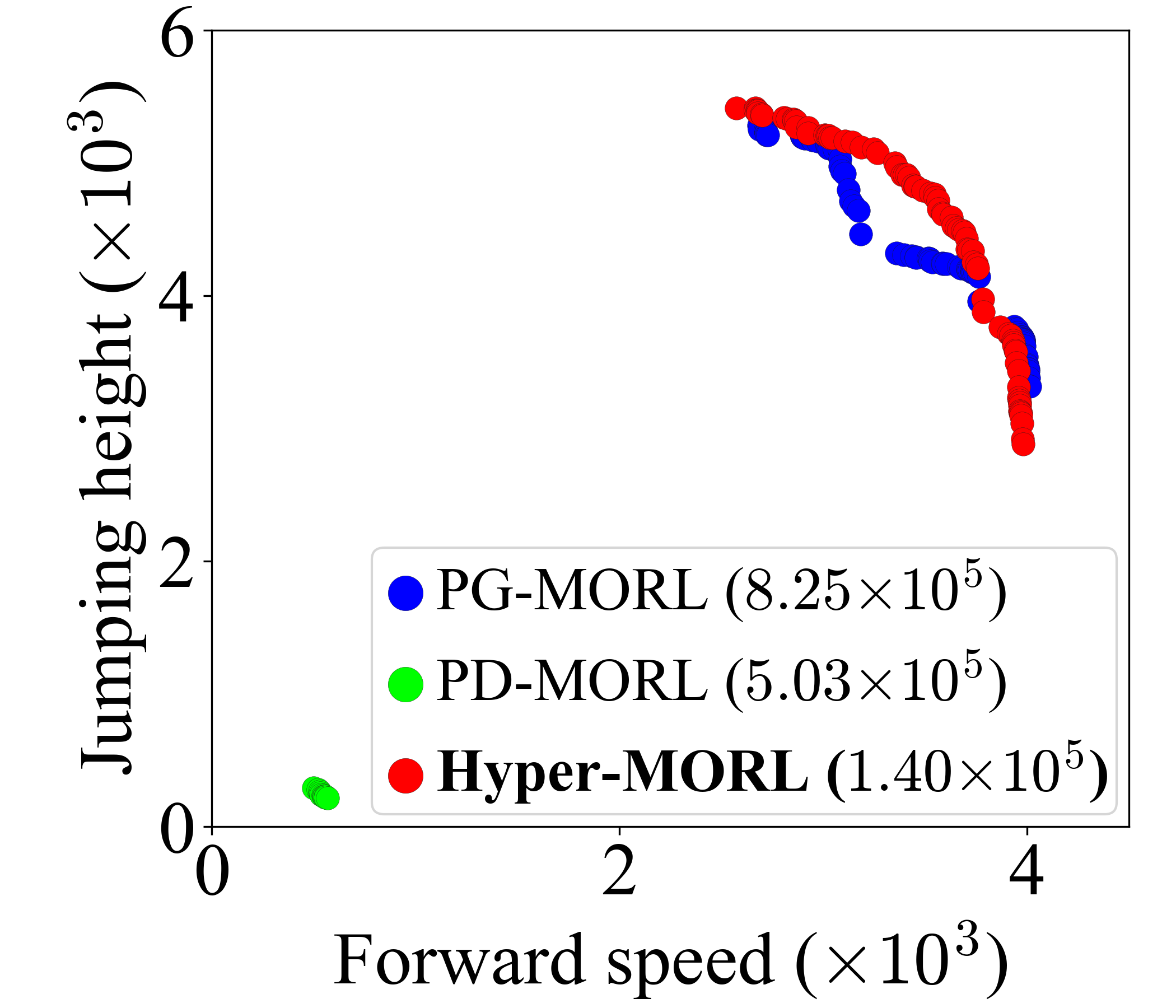}
    \caption{MO-Hopper-v2}
\end{subfigure}
\begin{subfigure}[c]{0.24\textwidth}
    \includegraphics[width=\textwidth]{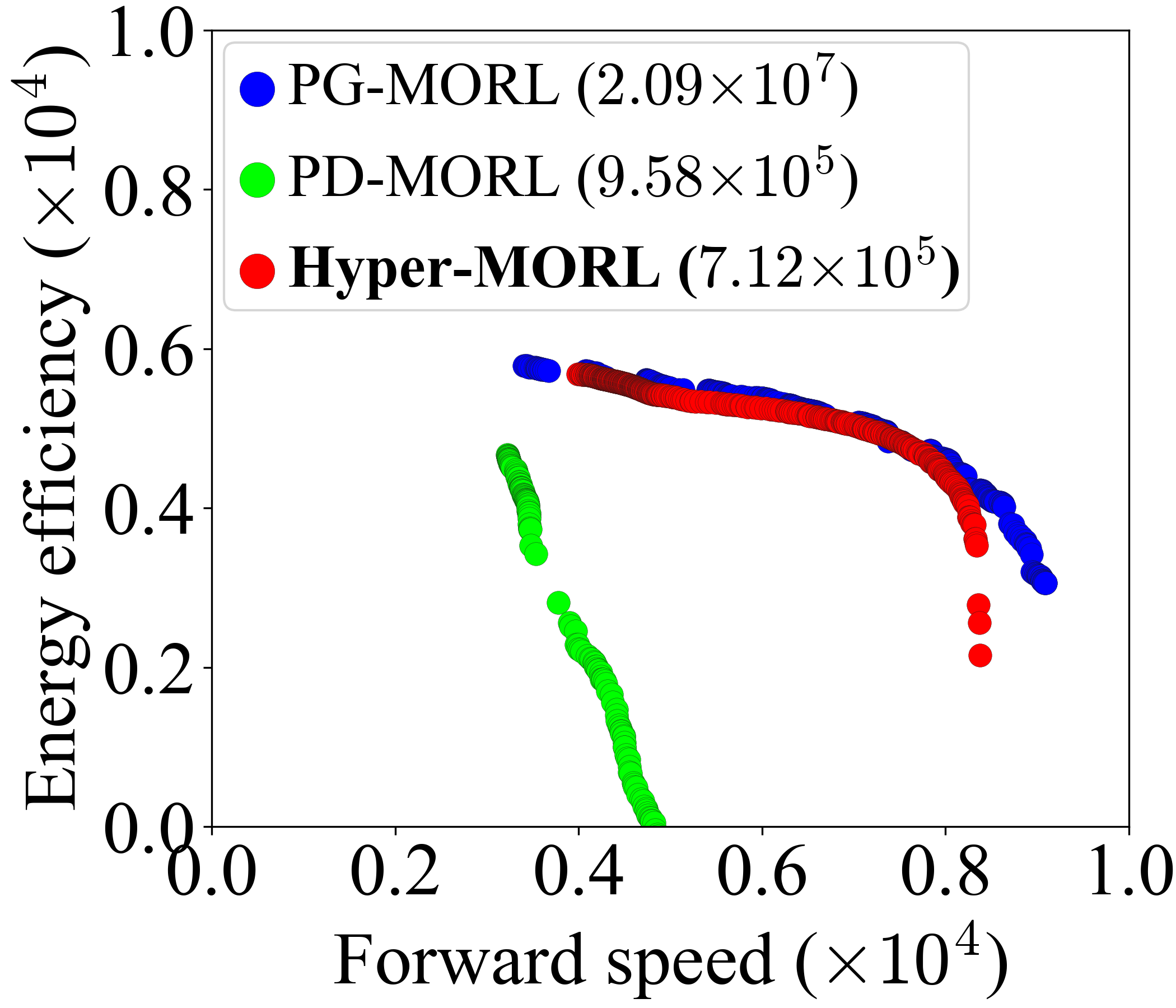}
    \caption{MO-Huamnoid-v2}
\end{subfigure}
\begin{subfigure}[c]{0.48\textwidth}
\centering
    \includegraphics[width=\textwidth]{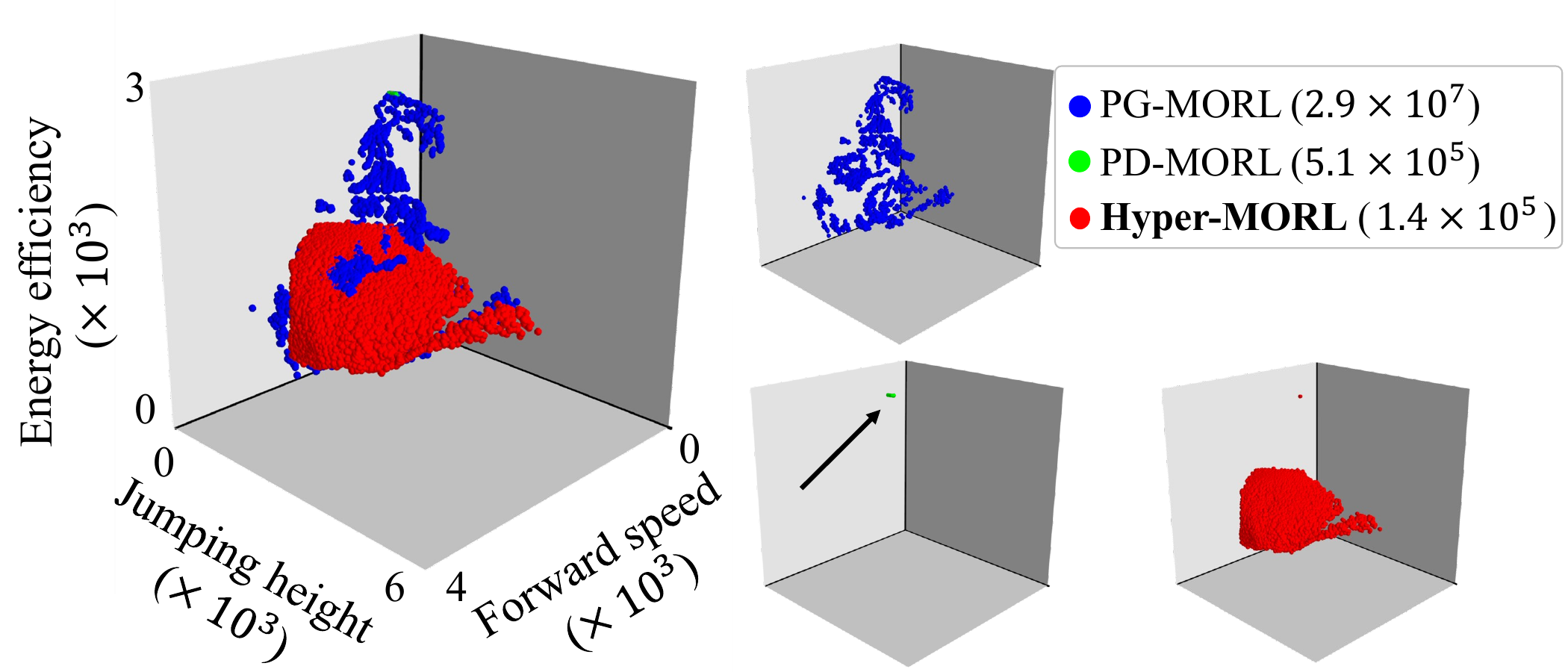}
    \caption{MO-Hopper-v3}
\end{subfigure}
    \caption{Visualization of all Pareto-optimal policies obtained by each MORL algorithm on each of six two-objective problems (a)-(f) and one three-objective problem (g). The total number of required parameters to represent these policies is shown in the parenthesis for each algorithm. A specific run with the median HV value among nine runs is shown for each algorithm.}
    \label{fig:visualization_2obj}
\end{figure*}
\subsection{Performance Comparison}\label{sec:performance_cp}
We first compare the quality of the policy sets learned by the three algorithms. A widely-used indicator hypervolume (HV)~\cite{zitzler2003performance,shang2020survey} is used for the quality assessment, which is explained in Appendix E. For the evolutionary learning algorithm (i.e., PG-MORL), we evaluate the final population with population sizes 200 and 420 for 2-objective and 3-objective problems, respectively. For a fair comparison, we sample the same number of uniformly distributed preferences for each of the single model-based algorithms (i.e., PD-MORL and Hyper-MORL), and evaluate the output policies. Table~\ref{tab:HV} shows the HV results of each algorithm on the seven test problems. As we can see, Hyper-MORL has the best HV performance on 4 out of 7 problems and the worst only on MO-HalfCheetah-v2. Thus, Hyper-MORL has the best average rank in Table~\ref{tab:HV}.\par 
To test the ability of each algorithm to generate the continuous Pareto front, we use a single run with the median HV value for each algorithm. For PG-MORL, we plot all Pareto-optimal policies stored in the unbounded external archive. For Hyper-MORL and PD-MORL, we input a number of uniform preferences and plot all Pareto-optimal policies corresponding to these preferences. The number of input preferences is set as 2,000 and 200,100 for 2-objective and 3-objective problems, respectively. Figure~\ref{fig:visualization_2obj} shows all Pareto-optimal policies from each algorithm on each test problem. Except for MO-HalfCheetah-v2, Hyper-MORL generates a continuous approximation of the Pareto front with high quality for each problem. Especially for MO-Hopper-v3 in Figure~\ref{fig:visualization_2obj} (g), Hyper-MORL generates a much more dense approximation than the other two methods. As shown in parenthesis in Figure~\ref{fig:visualization_2obj}, Hyper-MORL has the smallest number of parameters for each problem among three algorithms.\par
\begin{figure}[!t]
    \centering
    \includegraphics[width=1.0\linewidth]{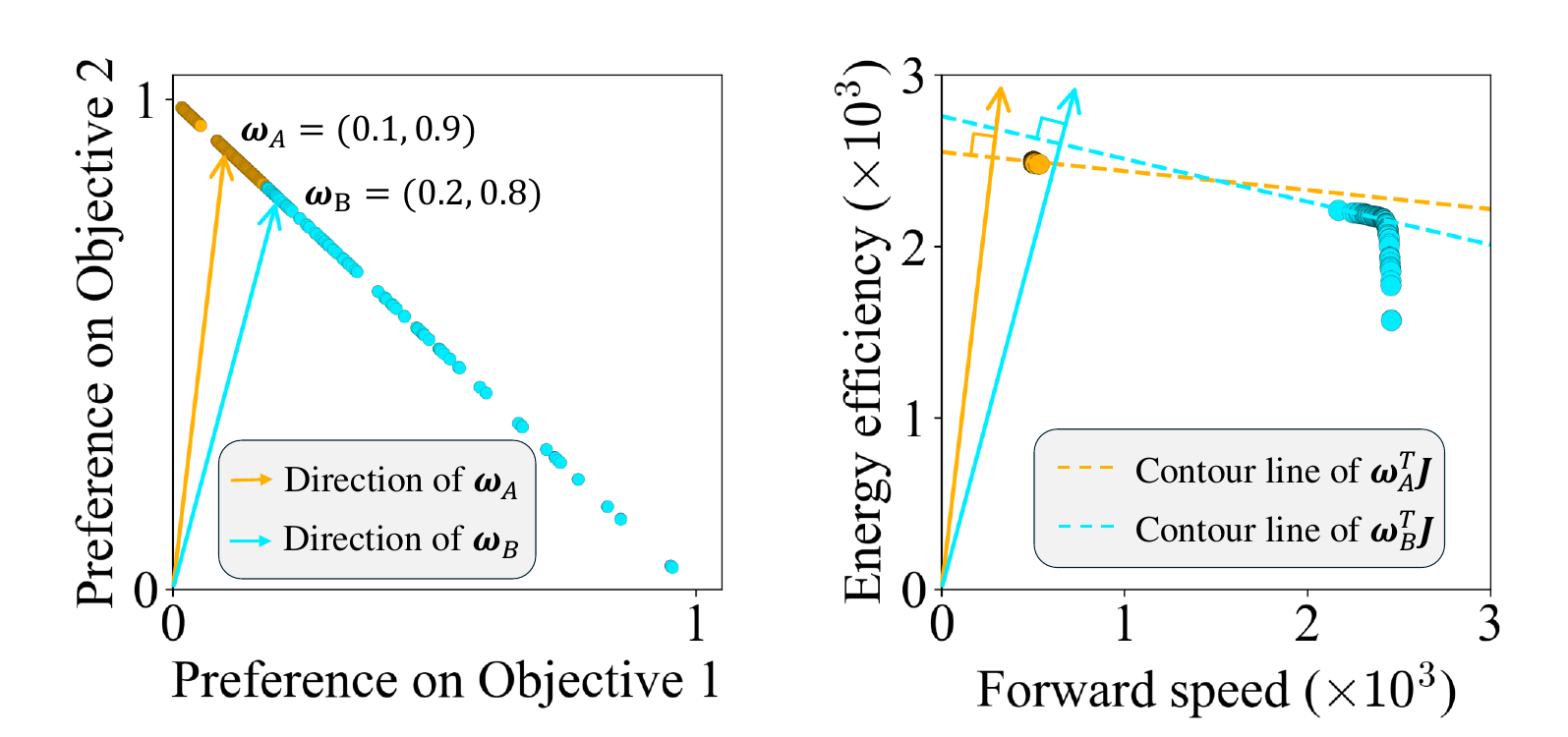}
    \caption{Explanation of the poor performance of Hyper-MORL on MO-HalfCheetah-v2 in Figure~\ref{fig:visualization_2obj} (b). The input preferences (Left) and their corresponding Pareto-optimal policies (Right) are plotted in two colors. The preferences in yellow (blue) correspond to the policies in yellow (blue). Two similar preferences $\bm{\omega}_{A}$ and $\bm{\omega}_{B}$ lead to clearly two different policies in the right figure.}
    \label{fig:poor_halfCheetah}
\end{figure}

 
Hyper-MORL shows poor performance only on MO-HalfCheetah-v2 in Figure~\ref{fig:visualization_2obj} (b). The reason is explained in Figure~\ref{fig:poor_halfCheetah}. As we can see, many preferences are mapped to the top left corner of the objective space by Hyper-MORL. This is because the linear scalarization function cannot find solutions on flat and concave regions of the Pareto front. The search for those regions by non-linear scalarization function is a future research topic.\par
\begin{figure}[!t]
    \centering
\includegraphics[width=0.8\linewidth]{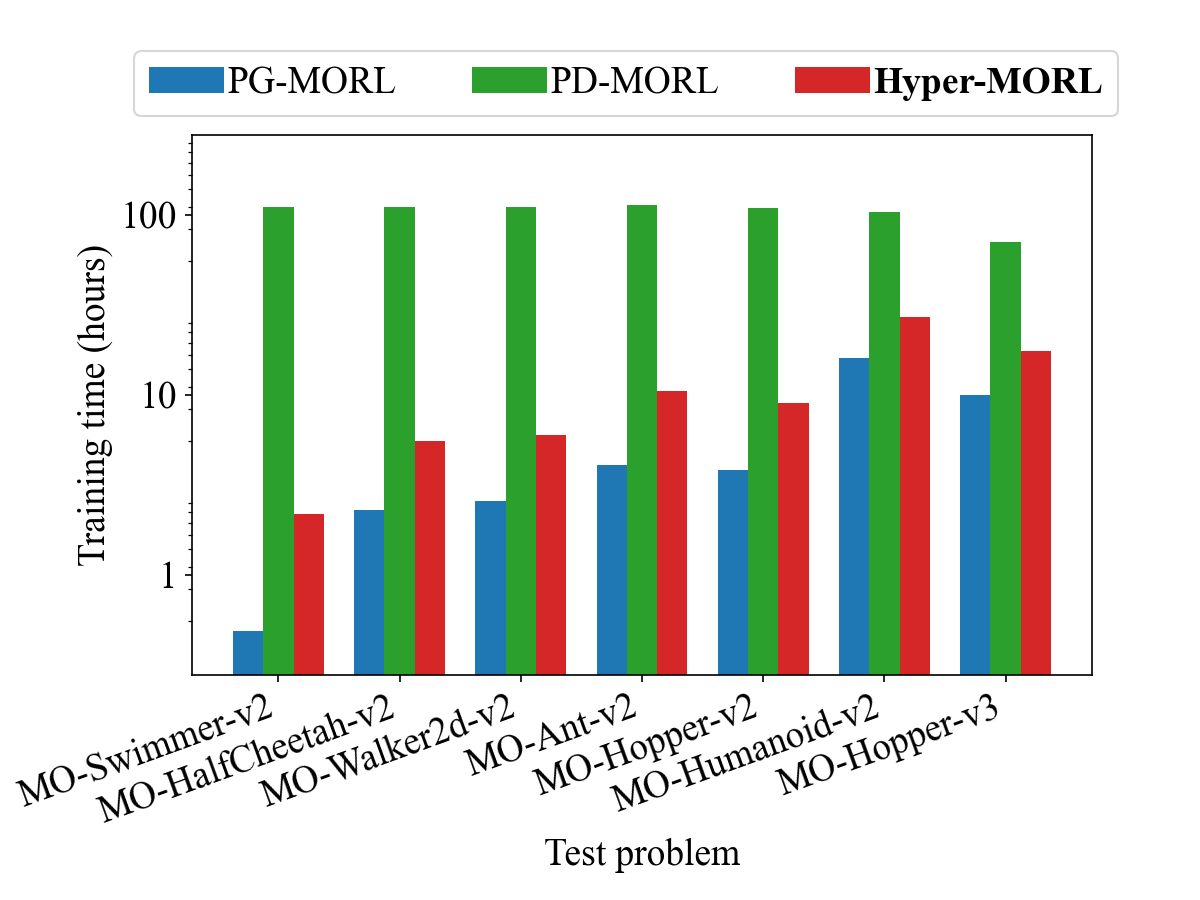}
    \caption{The average training time required for each algorithm on each test problem. The termination condition is shown in Table~\ref{tab:termination_condition}.}
    \label{fig:computation_time}
\end{figure}

From Table~\ref{tab:HV} and Figure~\ref{fig:visualization_2obj}, we can see that PD-MORL (green points in Figure~\ref{fig:visualization_2obj}) has poor performance on MO-Hopper-v2, MO-Humanoid-v2, and MO-Hopper-v3, which could be due to two reasons. One reason is that the performance of PD-MORL highly relies on an interpolation procedure with some key solutions, as mentioned in the original paper~\cite{basaklar2023pdmorl}. However, PD-MORL could not find any good key solutions for these problems. The other reason is that the number of environment steps (shown in Table~\ref{tab:termination_condition}) in PD-MORL may be insufficient. Figure~\ref{fig:computation_time} shows the training time of each algorithm. As we can see, PD-MORL needs the longest training time. For MO-Humanoid-v2, the training in PD-MORL for $1\times 10^{7}$ environment steps needs about four days, while the training in Hyper-MORL and PG-MORL for $1.2\times 10^{8}$ environment steps needs less than one day. If we use the same number of environment steps in all three algorithms, the training in PD-MORL will take about 40 days.\par
\section{Further Studies}\label{sec:sensitivity_analysis}
\subsection{Investigation of the Learned Pareto Set}\label{sec:investigation}
\begin{figure}
    \centering
    \begin{subfigure}[c]{0.32\linewidth}
    \includegraphics[width=\linewidth]{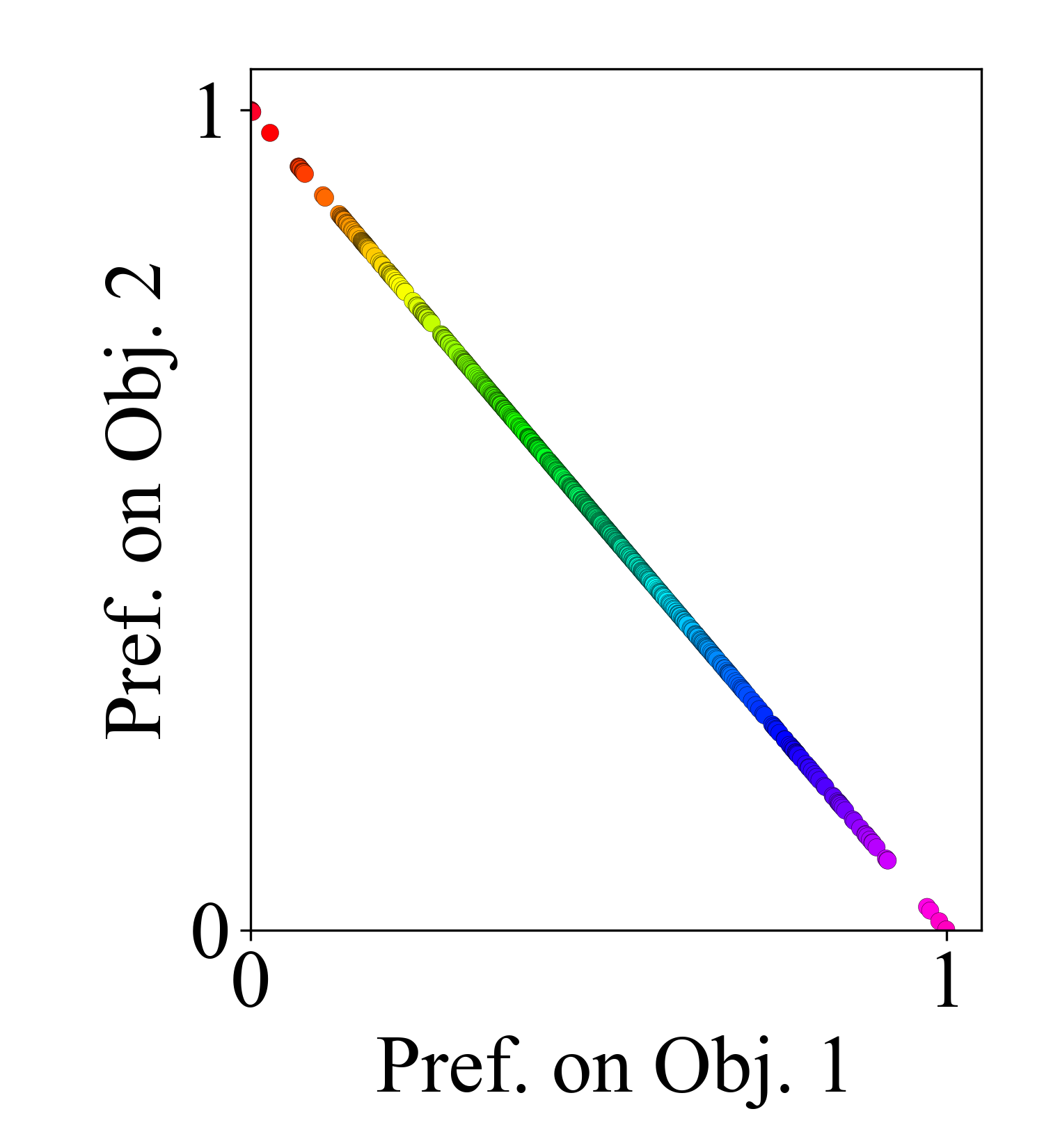}
    \caption{Preference space}
\end{subfigure}
  \begin{subfigure}[c]{0.32\linewidth}
    \includegraphics[width=\linewidth]{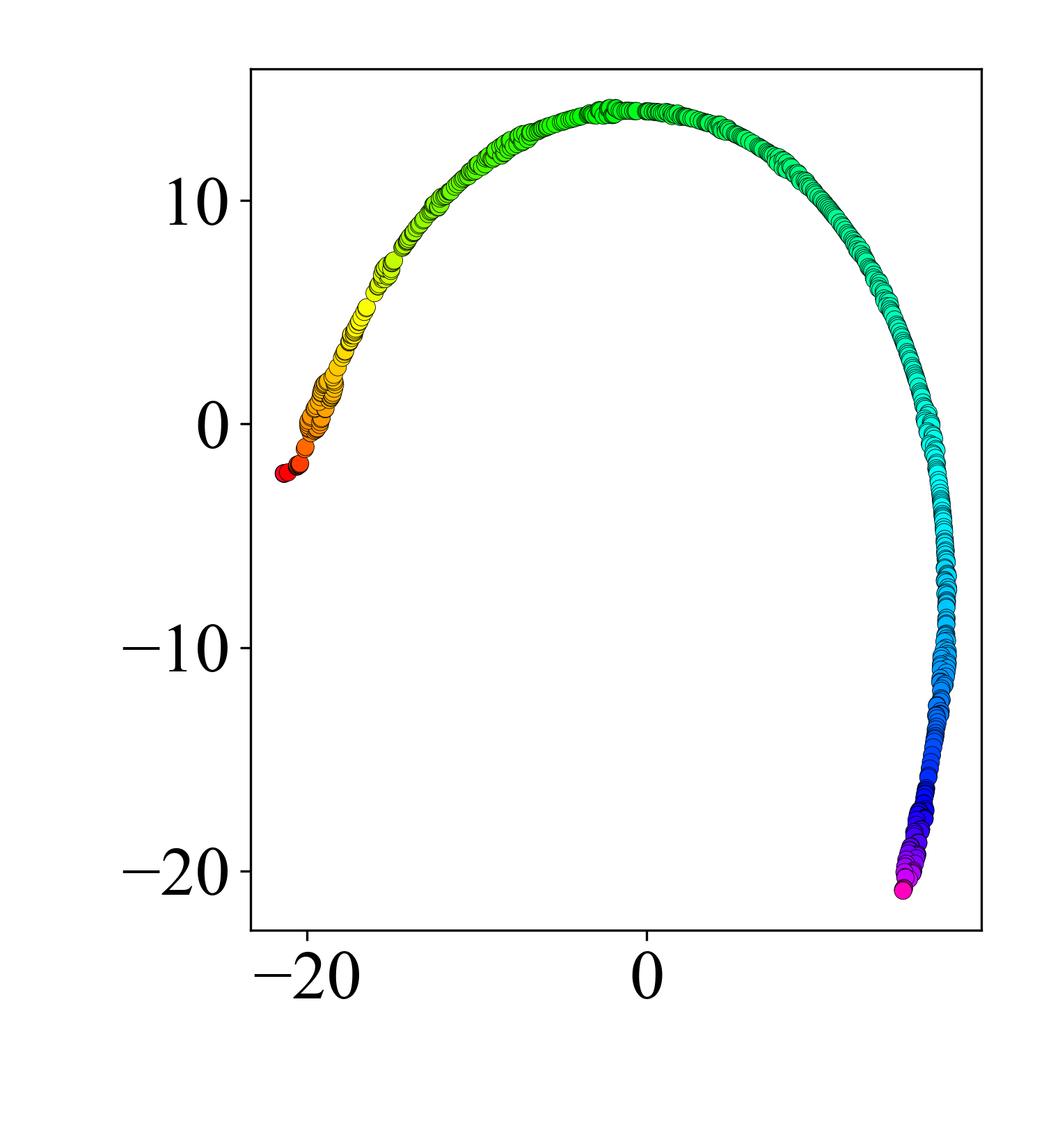}
    \caption{Parameter space}
\end{subfigure}
  \begin{subfigure}[c]{0.32\linewidth}
    \includegraphics[width=\linewidth]{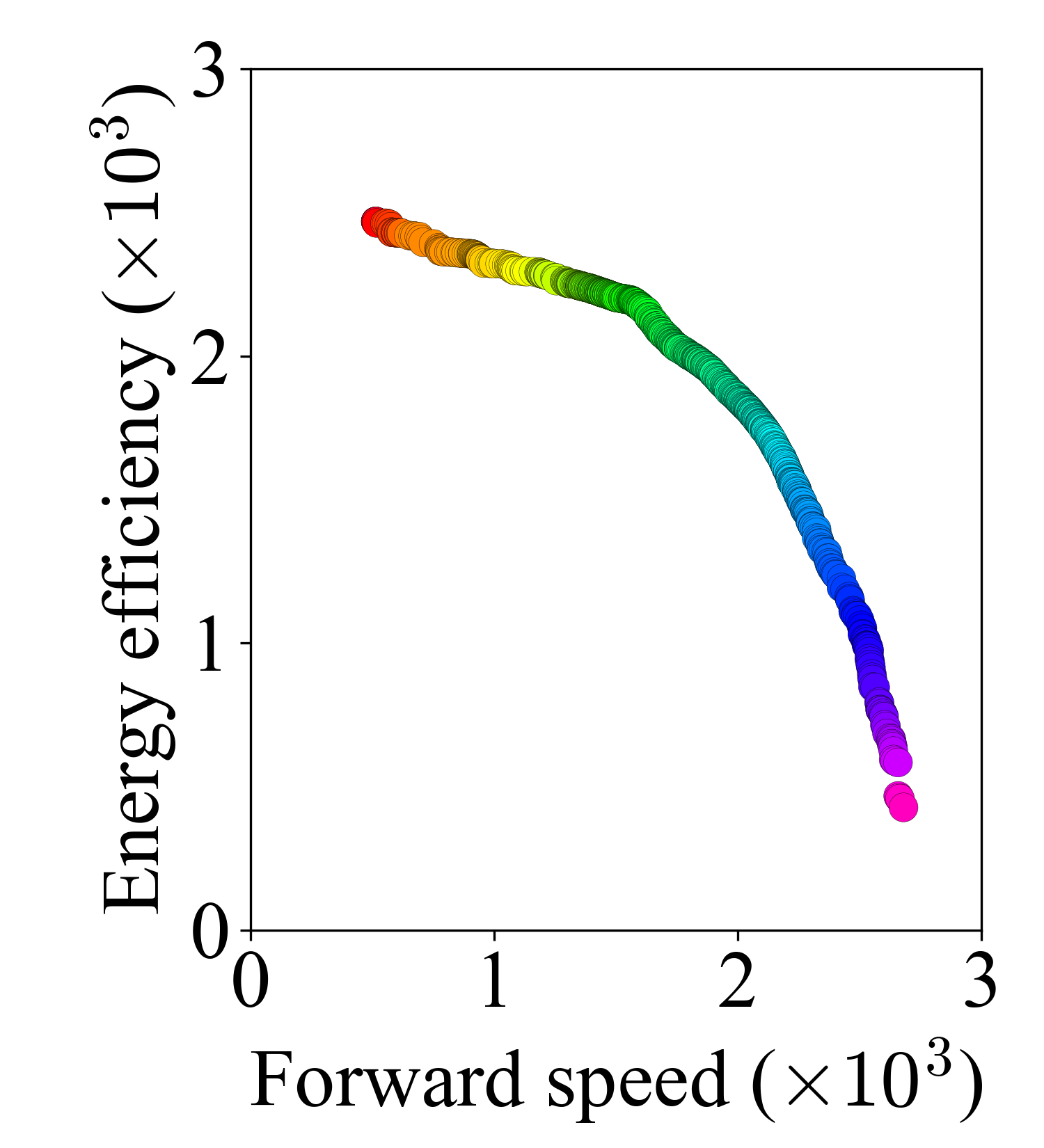}
    \caption{Objective space}
\end{subfigure}
\caption{Visualization of the Pareto-optimal policies obtained by Hyper-MORL for MO-Walker2d-v2 problem in the parameter space (b) and objective space (c), and their corresponding input to the hypernet in the preference space (a). Each policy is plotted with the same color as its corresponding input preference. t-SNE is used for visualization in the 11,150-dimensional parameter space.}
    \label{fig:mapping}
\end{figure}
We further investigate the relation between the input preferences and the Pareto sets learned by Hyper-MORL. For better visualization, each preference is plotted with a different color in Figure~\ref{fig:mapping} (a). Then, each policy in the learned Pareto set is plotted with the same color as its corresponding input preference in Figure~\ref{fig:mapping} (b) and (c) for MO-Walker2d-v2. Dominated policies and the related preferences are not shown in Figure~\ref{fig:mapping}. We show the learned Pareto set in both the parameter space and the objective space. Since the dimension of the parameter space is high, we use t-SNE~\cite{van2008visualizing} to visualize the high-dimensional parameter space in a two-dimensional space. Visualization results for the other problems are included in Appendix D.1. In Figure~\ref{fig:mapping}, we can observe a clear relation between the preferences in (a) and the Pareto front in (c). For example, the top-left solution in Figure 5 (c) with the largest second objective value is obtained from the preference (0, 1).\par
\begin{figure}[]
    \centering
\begin{subfigure}{0.48\linewidth}
    \includegraphics[width=\linewidth]{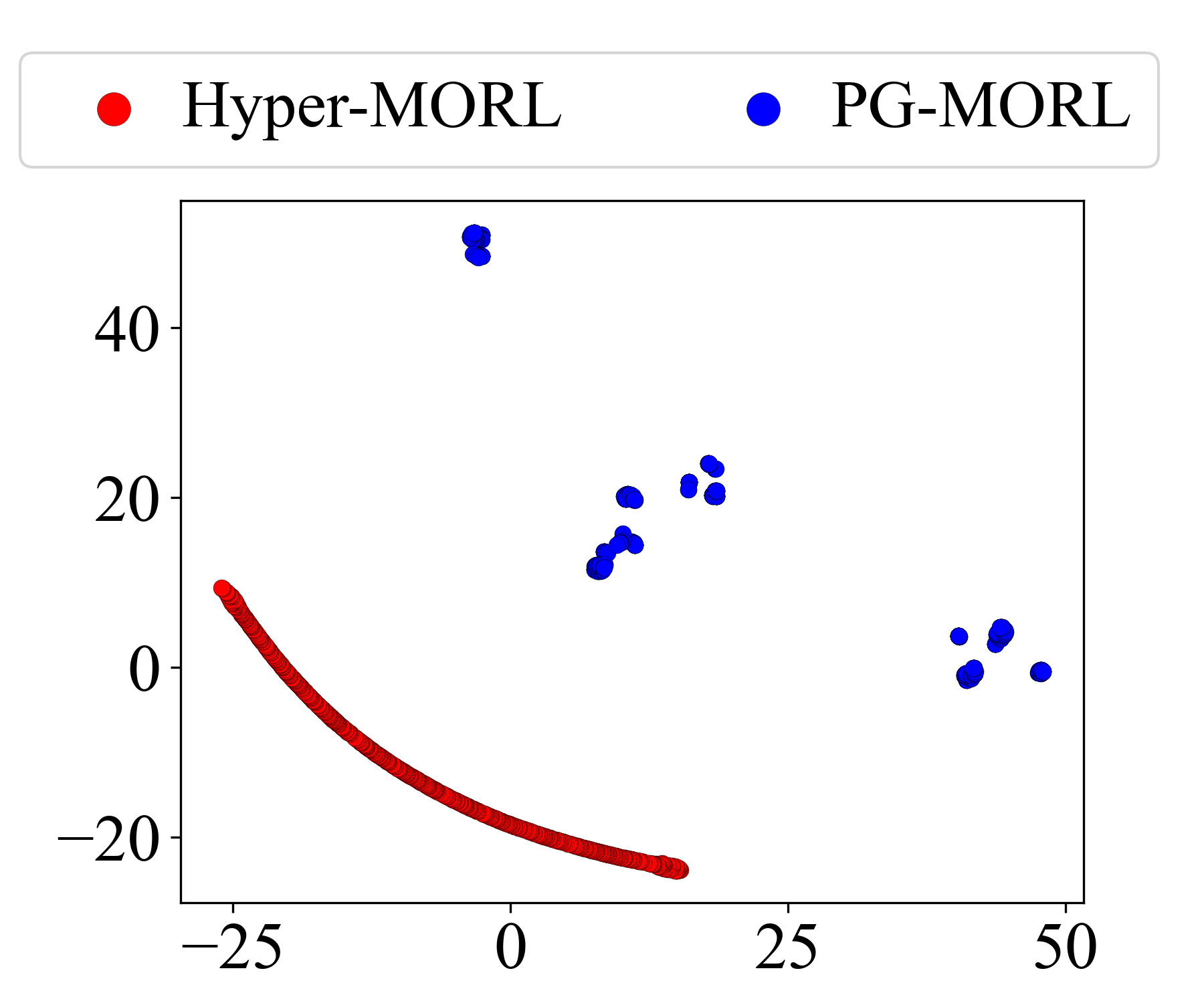}
    \caption{MO-Humanoid-v2}
\end{subfigure}
\begin{subfigure}{0.48\linewidth}
     \includegraphics[width=\linewidth]{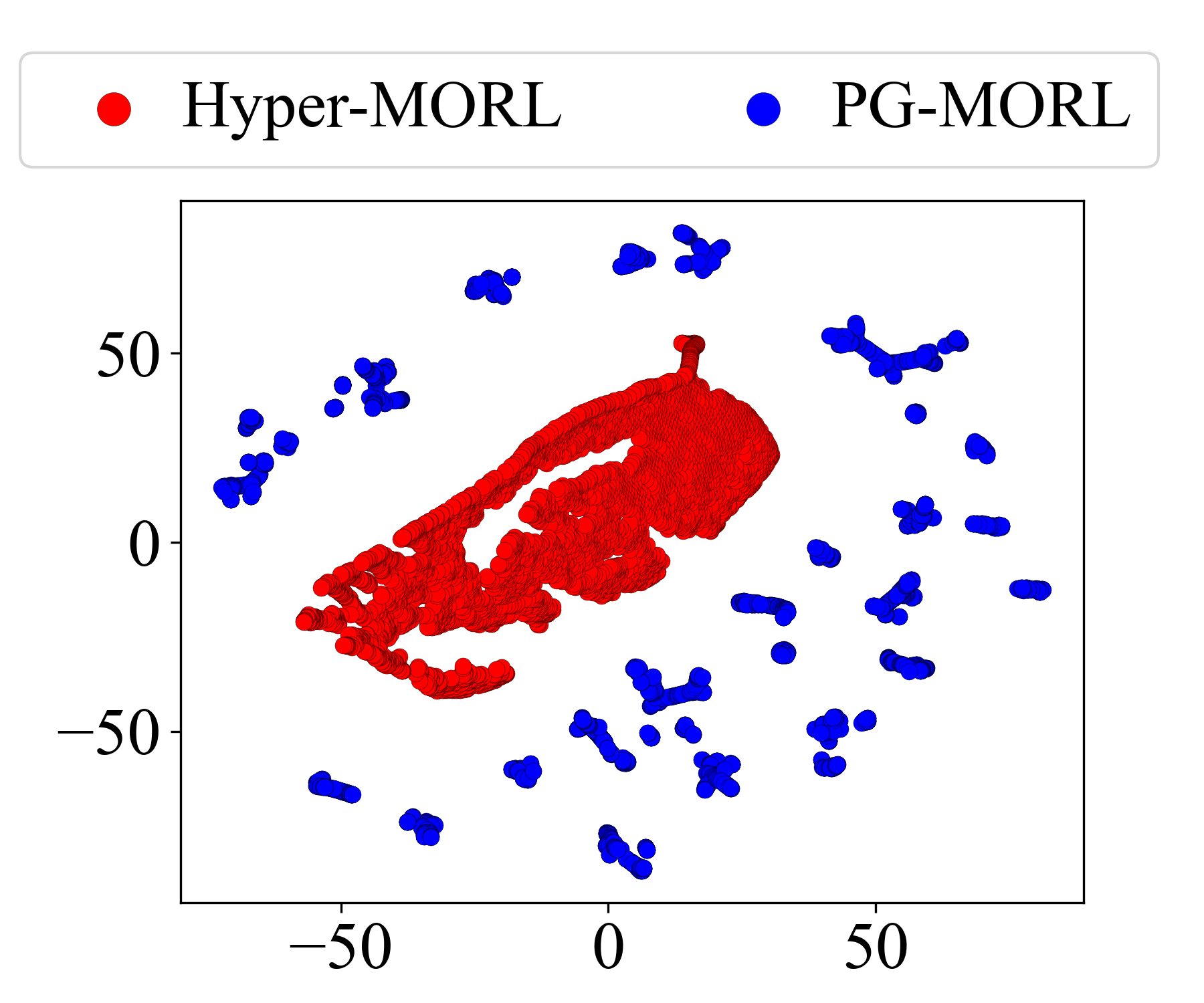}
     \caption{MO-Hopper-v3}
\end{subfigure}
    \caption{Comparison of the policy sets learned by Hyper-MORL and PG-MORL in the parameter space on MO-Humanoid-v2 and MO-Hopper-v3. Figure~\ref{fig:visualization_2obj} (f) and Figure~\ref{fig:visualization_2obj} (g) are the corresponding visualization in the objective space for (a) and (b).}
\label{fig:comparison_parameter_space}
\end{figure}

One interesting observation from Figure~\ref{fig:mapping} (b) is that the shape of the Pareto set learned by Hyper-MORL is a simple curve after the dimensionality reduction from the 11,150-dimensional parameter space. This observation is different from reported results on PG-MORL~\cite{xu2020prediction} where it was shown that a single continuous policy family cannot represent the Pareto set of these continuous robot control problems well. Thus, we further compare the two policy sets learned by Hyper-MORL and PG-MORL in the parameter space. PD-MORL is not considered since it has a different parameter space from Hyper-MORL and PG-MORL. Figure~\ref{fig:comparison_parameter_space} shows the comparison results on the two-objective MO-Humanoid-v2 and three-objective MO-Hopper-v3 problems. As we can see, Hyper-MORL learns a single continuous policy family in Figure~\ref{fig:comparison_parameter_space} (a) and closely related policy families in Figure~\ref{fig:comparison_parameter_space} (b) while PG-MORL learns a set of disjoint policy families.\par
\begin{figure}[!h]
    \centering
    \begin{subfigure}[c]{0.49\linewidth}
\includegraphics[width=\linewidth]
{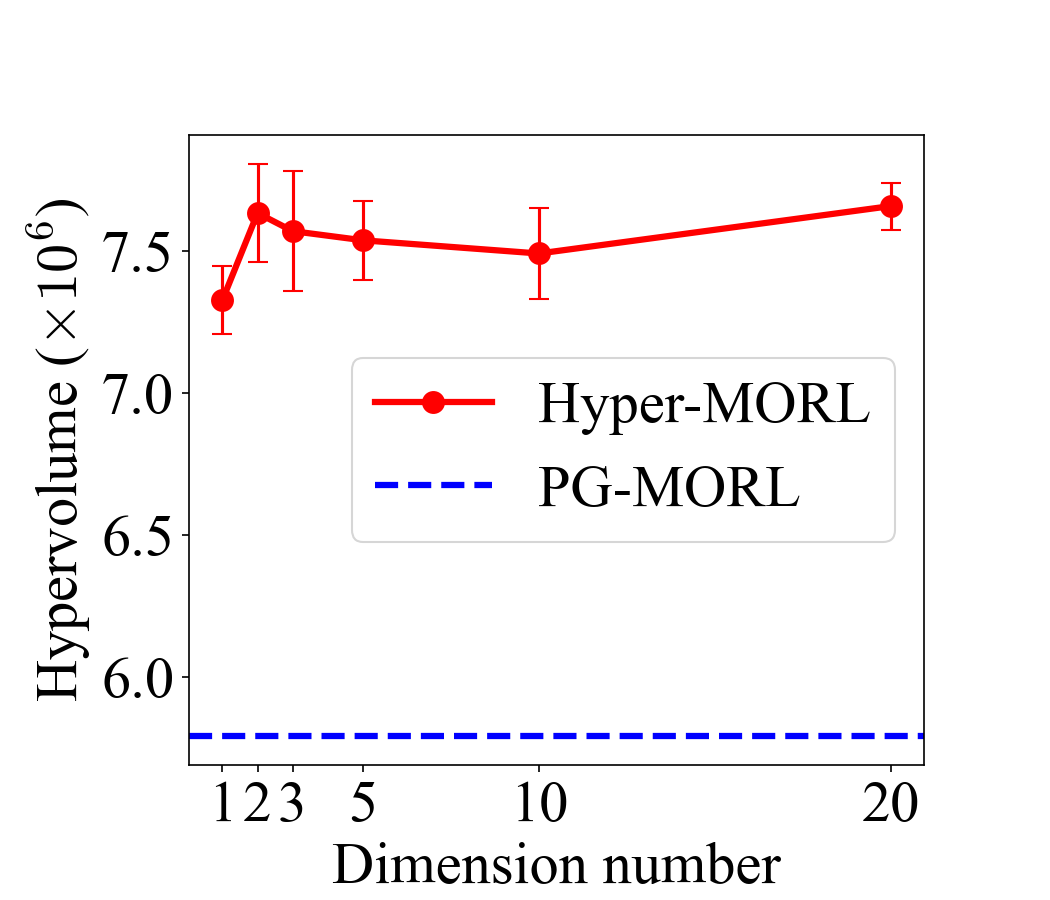}
        \caption{MO-Ant-v2}
    \end{subfigure}
    \begin{subfigure}[c]{0.49\linewidth}
        \includegraphics[width=\linewidth]
        {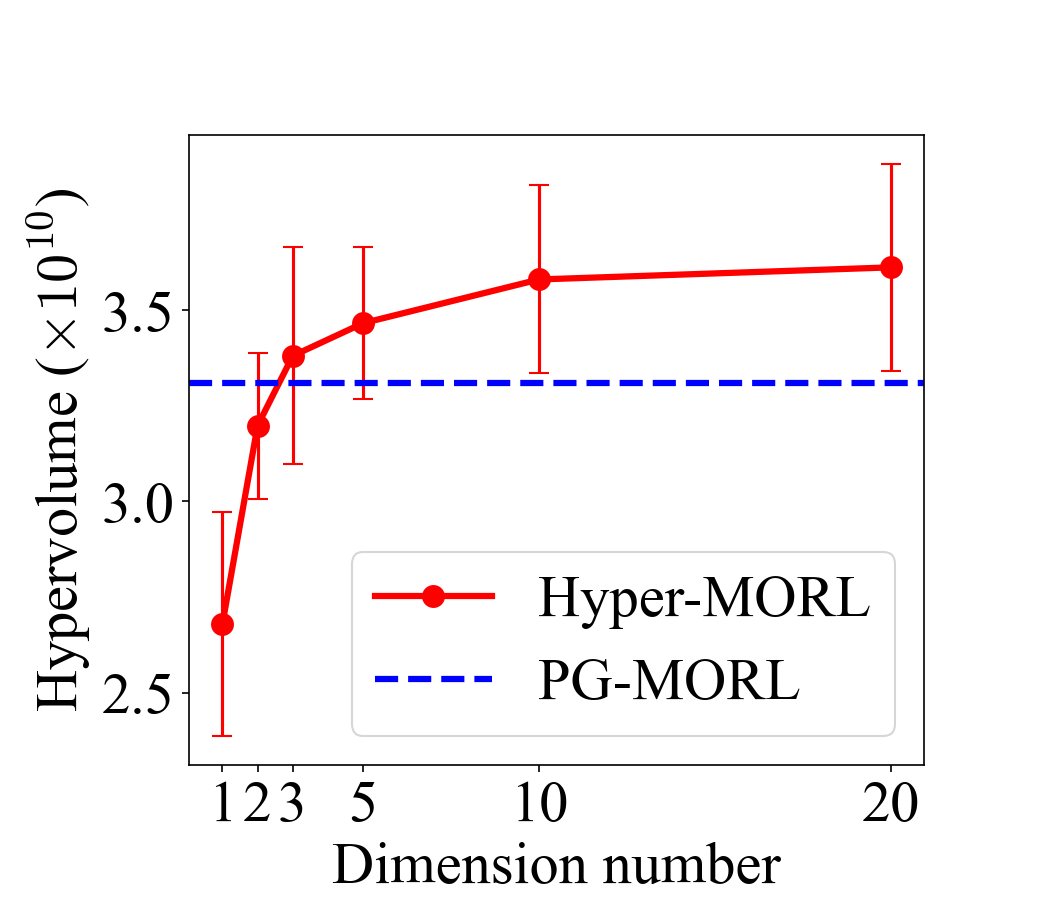}
        \caption{MO-Hopper-v3}
    \end{subfigure}
    \caption{Effects of the dimensionality $d$ of the reduced parameter space on the hypervolume performance of Hyper-MORL. The red vertical line shows the standard deviation among nine runs.}
    \label{fig:reduced_dimension}
\end{figure}

 As discussed in Section~\ref{sec:representation}, the Pareto set learned by Hyper-MORL is located in a $d$-dimensional parameter subspace where $d$ is a parameter for the hypernet used in Hyper-MORL. We examine the effects of $d$ on the hypervolume performance of Hyper-MORL using six values of $d$: $d=$1, 2, 3, 5, 10, 20. Figure~\ref{fig:reduced_dimension} shows the results of the two-objective MO-Ant-v2 and three-objective MO-Hopper-v3 problems. As a baseline, the hypervolume performance of PG-MORL is shown by the blue dotted line. In Figure~\ref{fig:reduced_dimension} (a), the hypervolume performance of Hyper-MORL is better than PG-MORL even when $d=1$. That means the Pareto set of MO-Ant-v2 is well approximated by Hyper-MORL with a curved line in the parameter space. For the three-objective MO-Hopper-v3 problem in Figure~\ref{fig:reduced_dimension} (b), good results are obtained when $d>2$. The results for the other problems are included in Appendix D.2. Our experimental results show that the Pareto sets of most problems are well approximated by Hyper-MORL in low-dimensional parameter subspace.\par
\subsection{Effects of the Warm-Up Stage}
\begin{figure}[!h]
    \centering
\includegraphics[width=0.8\linewidth]{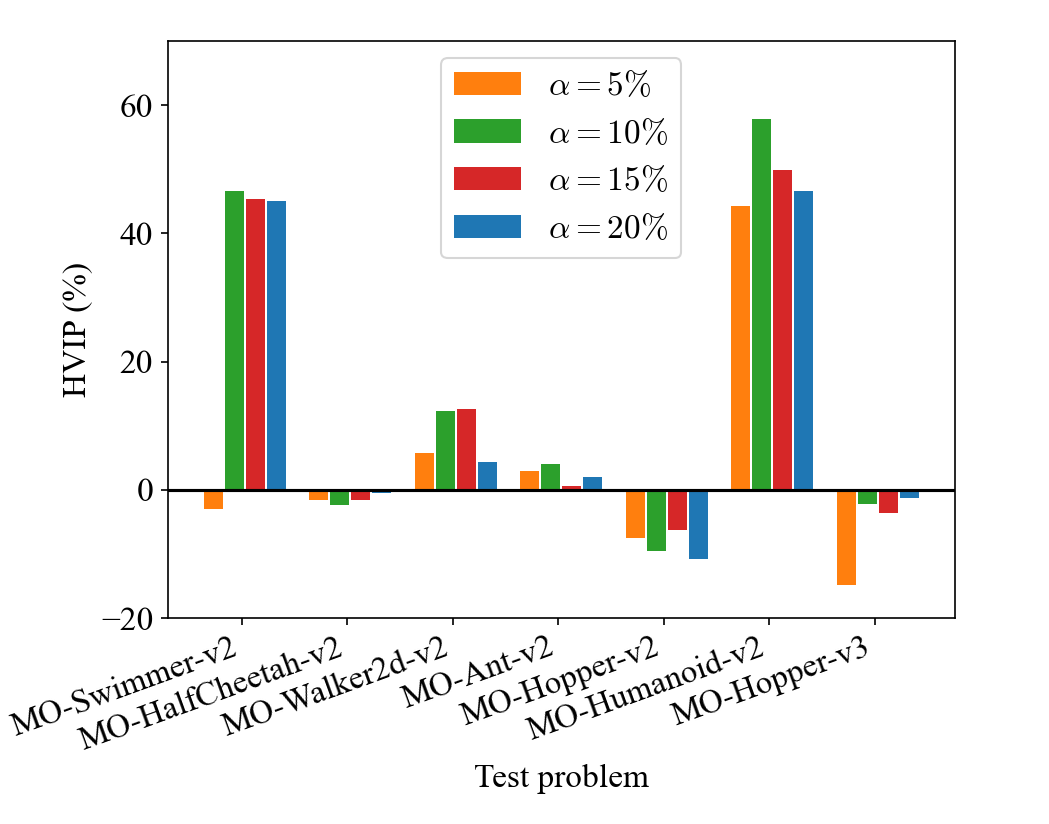}
    \caption{The hypervolume improvement percentage (HVIP) of Hyper-MORL by the warm-up stage with different $\alpha$ values.}
    \label{fig:warm_up}
\end{figure}

To examine the effects of the warm-up stage, we test four specifications of $\alpha$: $0\%$, $5\%$, $10\%$, and $20\%$. $\alpha=0\%$ means that the warm-up stage is not performed, and the hypernet is initialized by the Bias-HyperInit~\cite{jacob2023hypernetworks}. All the other settings except for $\alpha$ are the same as in the previous experiments. For each specification (i.e., $\alpha=x\%$), we calculate the hypervolume improvement percentage (HVIP) from the results with $\alpha=0\%$ as follows:
\begin{equation}
\text{HVIP}=\frac{\text{HV}_{\alpha=x\%}-\text{HV}_{\alpha=0\%}}{\text{HV}_{\alpha=0\%}}\times 100.
\end{equation}
Here positive and negative HVIP values mean positive and negative effects of the warm-up stage with $\alpha=x\%$. The HVIP results are shown in Figure~\ref{fig:warm_up}. We can see that the warm-up stage has a large positive effect on the hypervolume performance for MO-Swimmer and MO-Humanoid-v2 (i.e., more than $40\%$ improvement). For the other problems, the positive and negative effects of the warm-up stage are small (less than $10\%$). Figure~\ref{fig:warm_up} also shows that too small value (e.g., $5\%$) or too large value (e.g., $20\%$) for the parameter $\alpha$ is not good. Our recommended value for $\alpha$ is $15\%$.\par
\section{Conclusions}\label{sec:conclusion}
In this paper, we proposed an MORL algorithm called Hyper-MORL, which learns a hypernet to approximate the Pareto set in a high-dimensional parameter space of deep policies by using a low-dimensional reduced parameter space. We compared Hyper-MORL with two state-of-the-art algorithms on seven multi-objective robot control problems. Our results showed that Hyper-MORL has the best overall performance and the least training parameters among compared algorithms. By the investigation of the Pareto set learned by Hyper-MORL, we found that the Pareto set of these multi-objective objective robot control problems can be well approximated by a manifold in a low-dimensional (e.g., 3-dimensional) subspace of the original high-dimensional (e.g., 11,150-dimensional) parameter space. Our results provide valuable insights for researchers to design new MORL algorithms. In the future, we try to combine a non-linear scalarization function into our method.

\appendix
\section*{Acknowledgments}
This work was supported by National Natural Science Foundation of China (Grant No. 62250710163, 62376115), Guangdong Provincial Key Laboratory (Grant No. 2020B121201001).
\bibliographystyle{named}
\bibliography{ijcai24,newRef}

\begin{thebibliography}{}

\bibitem[\protect\citeauthoryear{Abels \bgroup \em et al.\egroup }{2019}]{abels2019dynamic}
Axel Abels, Diederik Roijers, Tom Lenaerts, Ann Now{\'e}, and Denis Steckelmacher.
\newblock Dynamic weights in multi-objective deep reinforcement learning.
\newblock In {\em International Conference on Machine Learning}, volume~97, pages 11--20, Jun 2019.

\bibitem[\protect\citeauthoryear{Alegre \bgroup \em et al.\egroup }{2023}]{lucas2023sample}
Lucas~N. Alegre, Ana L.~C. Bazzan, Diederik~M. Roijers, Ann Now\'{e}, and Bruno~C. da~Silva.
\newblock Sample-efficient multi-objective learning via generalized policy improvement prioritization.
\newblock In {\em Autonomous Agents and Multi-Agent Systems}, page 2003–2012, 2023.

\bibitem[\protect\citeauthoryear{Barreto \bgroup \em et al.\egroup }{2020}]{Barreto2020fast}
André Barreto, Shaobo Hou, Diana Borsa, David Silver, and Doina Precup.
\newblock Fast reinforcement learning with generalized policy updates.
\newblock In {\em Proceedings of the National Academy of Sciences}, 2020.

\bibitem[\protect\citeauthoryear{Basaklar \bgroup \em et al.\egroup }{2023}]{basaklar2023pdmorl}
Toygun Basaklar, Suat Gumussoy, and Umit Ogras.
\newblock {PD}-{MORL}: Preference-driven multi-objective reinforcement learning algorithm.
\newblock In {\em International Conference on Learning Representations}, 2023.

\bibitem[\protect\citeauthoryear{Beck \bgroup \em et al.\egroup }{2023}]{jacob2023hypernetworks}
Jacob Beck, Matthew~Thomas Jackson, Risto Vuorio, and Shimon Whiteson.
\newblock Hypernetworks in meta-reinforcement learning.
\newblock In {\em Proceedings of The 6th Conference on Robot Learning}, volume 205, pages 1478--1487, 2023.

\bibitem[\protect\citeauthoryear{Chauhan \bgroup \em et al.\egroup }{2023}]{chauhan2023brief}
Vinod~Kumar Chauhan, Jiandong Zhou, Ping Lu, Soheila Molaei, and David~A Clifton.
\newblock A brief review of hypernetworks in deep learning.
\newblock {\em arXiv preprint arXiv:2306.06955}, 2023.

\bibitem[\protect\citeauthoryear{Chen \bgroup \em et al.\egroup }{2019}]{chen2019meta}
Xi~Chen, Ali Ghadirzadeh, M{\aa}rten Bj{\"o}rkman, and Patric Jensfelt.
\newblock Meta-learning for multi-objective reinforcement learning.
\newblock In {\em IEEE/RSJ International Conference on Intelligent Robots and Systems (IROS)}, pages 977--983. IEEE, 2019.

\bibitem[\protect\citeauthoryear{Galanti and Wolf}{2020}]{galanti2020on}
Tomer Galanti and Lior Wolf.
\newblock On the modularity of hypernetworks.
\newblock In {\em Advances in Neural Information Processing Systems}, volume~33, pages 10409--10419, 2020.

\bibitem[\protect\citeauthoryear{Hayes \bgroup \em et al.\egroup }{2022}]{hayes2022practical}
Conor~F. Hayes, Roxana R{\u{a}}dulescu, Eugenio Bargiacchi, Johan K{\"a}llstr{\"o}m, Matthew Macfarlane, Mathieu Reymond, Timothy Verstraeten, Luisa~M. Zintgraf, Richard Dazeley, Fredrik Heintz, et~al.
\newblock A practical guide to multi-objective reinforcement learning and planning.
\newblock {\em Autonomous Agents and Multi-Agent Systems}, 36(1):26, 2022.

\bibitem[\protect\citeauthoryear{Hillermeier}{2001}]{hillermeier2001nonlinear}
Claus Hillermeier.
\newblock {\em Nonlinear multiobjective optimization: a generalized homotopy approach}, volume 135.
\newblock Springer Science \& Business Media, 2001.

\bibitem[\protect\citeauthoryear{Huang \bgroup \em et al.\egroup }{2021}]{huang2021continual}
Yizhou Huang, Kevin Xie, Homanga Bharadhwaj, and Florian Shkurti.
\newblock Continual model-based reinforcement learning with hypernetworks.
\newblock In {\em IEEE International Conference on Robotics and Automation (ICRA)}, pages 799--805, 2021.

\bibitem[\protect\citeauthoryear{Lin \bgroup \em et al.\egroup }{2019}]{lin2019pareto}
Xi~Lin, Hui-Ling Zhen, Zhenhua Li, Qing-Fu Zhang, and Sam Kwong.
\newblock Pareto multi-task learning.
\newblock In {\em Advances in Neural Information Processing Systems}, volume~32, 2019.

\bibitem[\protect\citeauthoryear{Lin \bgroup \em et al.\egroup }{2022}]{lin2022pareto_combinatorial}
Xi~Lin, Zhiyuan Yang, and Qingfu Zhang.
\newblock Pareto set learning for neural multi-objective combinatorial optimization.
\newblock In {\em International Conference on Learning Representations}, 2022.

\bibitem[\protect\citeauthoryear{Lu \bgroup \em et al.\egroup }{2023}]{lu2023multiobjective}
Haoye Lu, Daniel Herman, and Yaoliang Yu.
\newblock Multi-objective reinforcement learning: Convexity, stationarity and {Pareto} optimality.
\newblock In {\em International Conference on Learning Representations}, 2023.

\bibitem[\protect\citeauthoryear{Navon \bgroup \em et al.\egroup }{2021}]{navon2021learning}
Aviv Navon, Aviv Shamsian, Ethan Fetaya, and Gal Chechik.
\newblock Learning the pareto front with hypernetworks.
\newblock In {\em International Conference on Learning Representations}, 2021.

\bibitem[\protect\citeauthoryear{Owen and Zhou}{2000}]{owen2000safe}
Art Owen and Yi~Zhou.
\newblock Safe and effective importance sampling.
\newblock {\em Journal of the American Statistical Association}, 95(449):135--143, 2000.

\bibitem[\protect\citeauthoryear{Parisi \bgroup \em et al.\egroup }{2014}]{parisi2014Policy}
Simone Parisi, Matteo Pirotta, Nicola Smacchia, Luca Bascetta, and Marcello Restelli.
\newblock Policy gradient approaches for multi-objective sequential decision making.
\newblock In {\em International Joint Conference on Neural Networks}, pages 2323--2330, 2014.

\bibitem[\protect\citeauthoryear{Parisi \bgroup \em et al.\egroup }{2017}]{parisi2017manifold}
Simone Parisi, Matteo Pirotta, and Jan Peters.
\newblock Manifold-based multi-objective policy search with sample reuse.
\newblock {\em Neurocomputing}, 263:3--14, 2017.

\bibitem[\protect\citeauthoryear{Pirotta \bgroup \em et al.\egroup }{2015}]{pirotta2015multi}
Matteo Pirotta, Simone Parisi, and Marcello Restelli.
\newblock Multi-objective reinforcement learning with continuous pareto frontier approximation.
\newblock In {\em Proceedings of the AAAI Conference on Artificial Intelligence}, volume~29, 2015.

\bibitem[\protect\citeauthoryear{Rame \bgroup \em et al.\egroup }{2023}]{rame2023rewarded}
Alexandre Rame, Guillaume Couairon, Corentin Dancette, Jean-Baptiste Gaya, Mustafa Shukor, Laure Soulier, and Matthieu Cord.
\newblock Rewarded soups: towards pareto-optimal alignment by interpolating weights fine-tuned on diverse rewards.
\newblock In {\em Advances in Neural Information Processing Systems}, 2023.

\bibitem[\protect\citeauthoryear{Rezaei-Shoshtari \bgroup \em et al.\egroup }{2023}]{2023rezaeihypernetworks}
Sahand Rezaei-Shoshtari, Charlotte Morissette, Francois~R. Hogan, Gregory Dudek, and David Meger.
\newblock Hypernetworks for zero-shot transfer in reinforcement learning.
\newblock In {\em Proceedings of the AAAI Conference on Artificial Intelligence}, volume~37, pages 9579--9587, Jun. 2023.

\bibitem[\protect\citeauthoryear{Roijers \bgroup \em et al.\egroup }{2013}]{roijers2013survey}
Diederik~M Roijers, Peter Vamplew, Shimon Whiteson, and Richard Dazeley.
\newblock A survey of multi-objective sequential decision-making.
\newblock {\em Journal of Artificial Intelligence Research}, 48:67--113, 2013.

\bibitem[\protect\citeauthoryear{Roijers \bgroup \em et al.\egroup }{2018}]{roijers2018multi}
Diederik~M Roijers, Denis Steckelmacher, and Ann Now{\'e}.
\newblock Multi-objective reinforcement learning for the expected utility of the return.
\newblock In {\em Proceedings of the Adaptive and Learning Agents Workshop at FAIM}, 2018.

\bibitem[\protect\citeauthoryear{Sarafian \bgroup \em et al.\egroup }{2021}]{sarafian2021Recomposing}
Elad Sarafian, Shai Keynan, and Sarit Kraus.
\newblock Recomposing the reinforcement learning building blocks with hypernetworks.
\newblock In {\em International Conference on Machine Learning}, volume 139, pages 9301--9312, 2021.

\bibitem[\protect\citeauthoryear{Schulman \bgroup \em et al.\egroup }{2017}]{schulman2017proximal}
John Schulman, Filip Wolski, Prafulla Dhariwal, Alec Radford, and Oleg Klimov.
\newblock Proximal policy optimization algorithms.
\newblock {\em arXiv preprint arXiv:1707.06347}, 2017.

\bibitem[\protect\citeauthoryear{Shang \bgroup \em et al.\egroup }{2021}]{shang2020survey}
Ke~Shang, Hisao Ishibuchi, Linjun He, and Lie~Meng Pang.
\newblock A survey on the hypervolume indicator in evolutionary multiobjective optimization.
\newblock {\em IEEE Transactions on Evolutionary Computation}, 25(1):1--20, 2021.

\bibitem[\protect\citeauthoryear{Van~der Maaten and Hinton}{2008}]{van2008visualizing}
Laurens Van~der Maaten and Geoffrey Hinton.
\newblock Visualizing data using {t-SNE}.
\newblock {\em Journal of Machine Learning Research}, 9(11), 2008.

\bibitem[\protect\citeauthoryear{Van~Moffaert and Now{\'e}}{2014}]{van2014multi}
Kristof Van~Moffaert and Ann Now{\'e}.
\newblock Multi-objective reinforcement learning using sets of pareto dominating policies.
\newblock {\em Journal of Machine Learning Research}, 15(1):3483--3512, 2014.

\bibitem[\protect\citeauthoryear{von Oswald \bgroup \em et al.\egroup }{2020}]{Oswald2020Continual}
Johannes von Oswald, Christian Henning, Benjamin~F. Grewe, and João Sacramento.
\newblock Continual learning with hypernetworks.
\newblock In {\em International Conference on Learning Representations}, 2020.

\bibitem[\protect\citeauthoryear{Xu \bgroup \em et al.\egroup }{2020}]{xu2020prediction}
Jie Xu, Yunsheng Tian, Pingchuan Ma, Daniela Rus, Shinjiro Sueda, and Wojciech Matusik.
\newblock Prediction-guided multi-objective reinforcement learning for continuous robot control.
\newblock In {\em International Conference on Machine Learning}, pages 10607--10616, 2020.

\bibitem[\protect\citeauthoryear{Yang \bgroup \em et al.\egroup }{2019}]{yang2019generalized}
Runzhe Yang, Xingyuan Sun, and Karthik Narasimhan.
\newblock A generalized algorithm for multi-objective reinforcement learning and policy adaptation.
\newblock In {\em Advances in Neural Information Processing Systems}, volume~32, 2019.

\bibitem[\protect\citeauthoryear{Zhang and Li}{2007}]{zhang2007moea}
Qingfu Zhang and Hui Li.
\newblock {MOEA/D}: A multiobjective evolutionary algorithm based on decomposition.
\newblock {\em IEEE Transactions on Evolutionary Computation}, 11(6):712--731, 2007.

\bibitem[\protect\citeauthoryear{Zitzler \bgroup \em et al.\egroup }{2003}]{zitzler2003performance}
Eckart Zitzler, Lothar Thiele, Marco Laumanns, Carlos~M Fonseca, and Viviane~Grunert Da~Fonseca.
\newblock Performance assessment of multiobjective optimizers: An analysis and review.
\newblock {\em IEEE Transactions on Evolutionary Computation}, 7(2):117--132, 2003.

\end{thebibliography}

\end{document}